\documentclass[10pt,twocolumn,letterpaper]{article}

\usepackage{cvpr}              % To produce the CAMERA-READY version

\usepackage{rotating}
\usepackage{multirow}

\usepackage[table]{xcolor} % permite \cellcolor en tablas
\usepackage{booktabs}      % mejora el aspecto de las tablas
\newcommand{\first}[1]{\cellcolor[HTML]{63C07A}#1}   % tono más oscuro
\newcommand{\second}[1]{\cellcolor[HTML]{BDDC8E}#1}  % tono intermedio
\newcommand{\third}[1]{\cellcolor[HTML]{EBE7A0}#1}            % tono más claro
\usepackage[most]{tcolorbox} % permite resaltado con bordes redondeados opcionales
\usepackage[dvipsnames, table]{xcolor}
\definecolor{firstColor}{HTML}{63C07A}
\definecolor{secondColor}{HTML}{BDDC8E}
\definecolor{thirdColor}{HTML}{EBE7A0}
\newcommand{\First}[1]{\colorbox{firstColor}{#1}}
\newcommand{\Second}[1]{\colorbox{secondColor}{#1}}
\newcommand{\Third}[1]{\colorbox{thirdColor}{#1}}

\definecolor{cvprblue}{rgb}{0.21,0.49,0.74}
\usepackage[pagebackref,breaklinks,colorlinks,allcolors=cvprblue]{hyperref}

\def\paperID{*****} % *** Enter the Paper ID here
\def\confName{CVPR}
\def\confYear{2026}

\title{SWNet: A Cross-Spectral Network for Camouflaged Weed Detection}

\author{Henry O. Velesaca$^{1,2}$, Luigi Miranda$^{1}$, Angel D. Sappa$^{1,3}$\vspace{2mm}\\
$^{1}$ESPOL Polytechnic University, Campus Gustavo Galindo, 090902, Guayaquil, Ecuador  \vspace{2mm}\\
$^2$Software Engineering Department, Research Center for Information and Communication \\ Technologies (CITIC-UGR), University of Granada, 18071, Granada, Spain \vspace{2mm}\\
$^3$Computer Vision Center, Universitat Aut\`onoma de Barcelona, 08193-Bellaterra, Barcelona, Spain \vspace{2mm}\\
{\tt\small \{hvelesac,luidamir\}@espol.edu.ec, sappa@ieee.org}
}

\begin{document}
\maketitle

\begin{abstract}
This paper presents SWNet, a bimodal end-to-end cross-spectral network specifically engineered for the detection of camouflaged weeds in dense agricultural environments. Plant camouflage, characterized by homochromatic blending where invasive species mimic the phenotypic traits of primary crops, poses a significant challenge for traditional computer vision systems. To overcome these limitations, SWNet utilizes a Pyramid Vision Transformer v2 backbone to capture long-range dependencies and a Bimodal Gated Fusion Module to dynamically integrate Visible and Near-Infrared information. By leveraging the physiological differences in chlorophyll reflectance captured in the NIR spectrum, the proposed architecture effectively discriminates targets that are otherwise indistinguishable in the visible range. Furthermore, an Edge-Aware Refinement module is employed to produce sharper object boundaries and reduce structural ambiguity. Experimental results on the Weeds-Banana dataset indicate that SWNet outperforms ten state-of-the-art methods. The study demonstrates that the integration of cross-spectral data and boundary-guided refinement is essential for high segmentation accuracy in complex crop canopies. The code is available on GitHub: \url{https://cod-espol.github.io/SWNet/}.
\end{abstract}

\section{Introduction}
\label{sec:intro}
For over 150 years, plant camouflage has been extensively studied as an evolutionary strategy employed by flora to defend against herbivores and predators \cite{niu2018plant}. This visual blending relies heavily on environmental illumination and the spectral properties of chlorophyll, which allow certain plants to match their surroundings seamlessly \cite{yang2024plantcamo}. In the context of modern precision agriculture, this natural phenomenon presents a significant challenge: invasive weeds often exhibit phenotypic characteristics—such as leaf shape, texture, and green coloration—that are nearly identical to those of the primary crops. This homochromatic blending effectively camouflages the weeds within the crop canopy, rendering traditional visual detection methods inadequate \cite{fan2024cross}.

Consequently, there is a pressing need to develop and apply Camouflaged Object Detection (COD) methodologies specifically tailored for agricultural environments. To overcome the limitations of the visible light spectrum, researchers increasingly rely on multimodal approaches that combine standard RGB images with Near-Infrared (NIR) data. NIR imaging is particularly effective in agriculture because it captures distinct physiological differences in cellular structure and chlorophyll reflectance that are imperceptible in standard RGB imagery \cite{sa2022deepnir}.

To address these challenges, this study evaluates the performance of state-of-the-art COD techniques applied to weed detection in dense crop environments. Specifically, we utilize the Weeds-Banana dataset to establish robust detection baselines. As presented in Table 1, metric evaluation results for each COD technique are reported for both the RGB and NIR baselines. Results are presented using the metric notation defined.

The manuscript is organized as follows. Section~\ref{sec:back} introduces related work, recent SOTA COD techniques, and methods that address the problem of the COD approach. Section~\ref{sec:prop} presents the proposed architecture. Then, Section~\ref{sec:exp} shows the experimental results using SOTA COD techniques and the proposed approach. Finally, conclusions is given in Section~\ref{sec:conclu}.

\section{Background}
\label{sec:back}
Automated weed detection has become a cornerstone of precision agriculture, significantly reducing the reliance on broad-spectrum herbicides. The evolution of visual recognition systems in this domain can be broadly categorized into classical computer vision methodologies and modern, deep learning-based camouflaged object detection frameworks.

\subsection{Classical Weed Detection Techniques}
Early approaches to weed detection relied heavily on handcrafted features and conventional image processing techniques. These methods primarily exploited the phenotypic differences between crops and weeds using morphological characteristics, texture descriptors, and color indices.

Also, traditional algorithms frequently utilize vegetation indices, such as the Excess Green (ExG) index, to separate plant matter from the soil background. Following segmentation, morphological operations and shape-based features (e.g., leaf area, perimeter, and aspect ratio) were extracted to classify the plants \cite{woebbecke1995color}.

In order to handle more complex canopies, researchers incorporated texture analysis methods like the Gray-Level Co-occurrence Matrix (GLCM) or Histogram of Oriented Gradients (HOG), coupled with classical machine learning classifiers such as Support Vector Machines (SVM) or Random Forests \cite{murad2023weed}.

While computationally efficient, these classical techniques exhibit significant degradation in performance under real-world field conditions. They are highly susceptible to variations in natural illumination, partial occlusions, and shadows. Most importantly, these methods fail when crops and weeds share highly similar spectral and morphological traits, a condition known as homochromatic blending, which renders traditional feature-extraction ineffective.

\subsection{Camouflaged Weed Detection Techniques}
To address the severe limitations of classical methods in dense agricultural environments, recent research has pivoted towards Camouflaged Object Detection (COD). Unlike standard object detection, COD is specifically designed to identify targets that are seamlessly embedded within their surroundings by mimicking the background's color, texture, and structural patterns \cite{fan2020camouflaged}.

Modern COD architectures leverage deep Convolutional Neural Networks (CNNs) and Vision Transformers (ViTs) to extract high-level semantic representations. Techniques such as boundary-guided attention, search-and-identification mechanisms, and multi-scale feature fusion have proven highly effective in distinguishing subtle boundary discrepancies between disguised weeds and the primary crop \cite{wang2024cross}.

In agricultural COD, the visual similarity between weeds and crops (such as in the Weeds-Banana dataset) is often too high for RGB data alone. Consequently, the integration of Near-Infrared (NIR) imagery has emerged as a crucial strategy. Since different plant species exhibit unique physiological responses and chlorophyll reflectance in the NIR spectrum, fusing RGB spatial details with NIR spectral signatures allows neural networks to "break" the visual camouflage that occurs in the visible spectrum \cite{sa2022deepnir}.

Despite these advancements, many existing COD networks struggle with the computational overhead required for real-time field deployment or fail to optimally fuse multi-modal data in highly cluttered crop canopies, highlighting the need for more specialized architectures.

\begin{table*}[t]
    \centering
    \caption{Distinctive characteristics of the evaluated SOTA COD techniques.}
    \begin{tabular}{l|cccclr}
        \hline
        \textbf{Technique} & \textbf{Source} & \textbf{Source} & \textbf{Year} & \textbf{Image Size} & \textbf{Backbone} & \textbf{\#Param.} \\
        & & \textbf{Type} & & \textbf{(px)} & & \textbf{(M)} \\
        \hline
        %BASNet \cite{qin2019basnet} & CVPR & Conference & 2019 & $256 \times 256$ & ResNet-34 \cite{he2016deep} & 87.06 \\
        SINet-v2 \cite{fan2021concealed} & TPAMI & Journal & 2021 & $352 \times 352$ & Res2Net-50 \cite{gao2019res2net} & 24.93 \\       
        BGNet \cite{chen2022boundary} & IJCAI & Conference & 2022 & $416 \times 416$ & Res2Net-50 \cite{gao2019res2net} & 77.80 \\
        C$^{2}$F-Net \cite{chen2022camouflaged} & TCSVT & Conference & 2022 & $352 \times 352$ & Res2Net-50 \cite{gao2019res2net} & 26.36 \\
        OCENet \cite{liu2022modeling} & WACV & Conference & 2022 & $352 \times 352$ & ResNet-50 \cite{he2016deep} & 58.17 \\
        EAMNet \cite{sun2023edge} & ICME & Conference & 2023 & $384 \times 384$ & Res2Net-50 \cite{gao2019res2net} & 30.51 \\
        DGNet \cite{ji2023deep} & MIR & Journal & 2023 & $352 \times 352$ & EfficientNet \cite{tan2019efficientnet} & 8.30 \\
        HitNet \cite{hu2023high} & AAAI & Conference & 2023 & $352 \times 352$ & PVTv2 \cite{wang2022pvt} & 25.73 \\
        %PCNet \cite{yang2024plantcamo} & arXiv & - & 2024 & $352 \times 352$ & PVTv2 \cite{wang2022pvt} & 27.66 \\
        %MRNet \cite{zhang2025hunt} & & 2025 & & &  \\        
        ARNet \cite{wang2025assisted} & ICMR & Conference & 2025 & $416 \times 416$ & SMT-Tiny \cite{lin2023scale} & 12.82\\
        CHNet \cite{wang2025efficient} & ICMR & Conference & 2025 & $416 \times 416$ & SMT-Tiny \cite{lin2023scale} & 11.20\\
        %CTF-Net \cite{zhang2025effective} & CVIU & Journal & 2025 & $384 \times 384$ & PVTv2 \cite{wang2022pvt} & 64.48 \\ 
        ARNet-v2 \cite{wang2025assistedv2} & arXiv & - & 2025 & $416 \times 416$ & Res2Net-50 \cite{gao2019res2net} & 34.12 \\
        SWNet (our) & CVPR & Conference & 2026 & $416 \times 416$ & PVTv2 \cite{wang2022pvt} & 42.32 \\
        
        \hline
    \end{tabular}
    \label{tab:networks_eval}
\end{table*}

\begin{table*}[!h]
    \centering
    \caption{Details of the training parameters used in evaluated SOTA COD techniques. Learning rate (LR); Batch size (BS).}
    \resizebox{2\columnwidth}{!}{
    \begin{tabular}{l|cccccc}
        \toprule
        Technique & Optimizer & LR & BS & Epochs & Scheduler & Loss function \\
        \midrule
        %BASNet \cite{qin2019basnet} & Adam & 1e-3 & 8 & 1000 & ReduceLROnPlateau & BCE + SSIM + IOU (multi-stage fusion) \\
        SINet-v2 \cite{fan2021concealed} & Adam & 1e-4 & 16 & 150 & Custom (Adjust LR) & Structure loss (weighted BCE + weighted IOU) \\       
        BGNet \cite{chen2022boundary} & Adam & 1e-4 & 12 & 100 & Custom (Poly LR) & Structure loss (weighted BCE + weighted IOU) + Dice loss (edge)\\
        C$^{2}$F-Net \cite{chen2022camouflaged} & AdaXW & 1e-4 & 32 & 50 & Custom (Poly LR) & Structure loss (weighted BCE + weighted IOU) \\
        OCENet \cite{liu2022modeling} & Adam & 1e-5 & 4 & 50 & StepLR & Uncertainty aware structure loss (weighted BCE + weighted IOU) \\
        EAMNet \cite{sun2023edge} & AdamW & 5e-5 & 16 & 150 & Custom (Adjust LR) & Hybrid loss (weighted BCE + weighted IOU) + Edge loss (edge) \\
        DGNet \cite{ji2023deep} & AdamW & 5e-5 & 16 & 150 & CosineAnnealingLR & Hybrid loss (weighted BCE + weighted IOU) + MSE loss (grad) \\
        HitNet \cite{hu2023high} & AdamW & 1e-4 & 8 & 150 & Custom (Adjust LR) & Structure loss (weighted BCE + weighted IOU) \\
        %PCNet \cite{yang2024plantcamo} & AdamW & 1e-4 & 8 & 150 & Custom (Adjust LR) & Structure loss (weighted BCE + weighted IOU) \\
        %CTF-Net \cite{zhang2025effective} & Adam & 1e-4 & 12 & 100 & Custom (Poly LR) & Structure loss (weighted BCE + weighted IOU) + Dice loss (edge)  \\ 
        ARNet \cite{wang2025assisted} & Adam & 5e-5 & 8 & 100 & StepLR &  Structure loss (wBCE + wIOU) + Edge loss\\
        CHNet \cite{wang2025efficient} & AdamW & 5e-5 & 8 & 100 & CosineAnnealingLR & Structure loss (wBCE + wIOU) \\
        ARNet-v2 \cite{wang2025assistedv2} & AdamW & 5e-5 & 4 & 100 & CosineAnnealingLR & Hybrid structure loss (wBCE + wIOU) \\
        SWNet (our) & AdamW & 1e-4 & 10 & 200 & CosineAnnealingLR & Structure loss (weighted BCE + weighted IOU) + Edge loss (edge)  \\
        \bottomrule
    \end{tabular}
    }
    \label{tab:networks_params}
\end{table*}

\begin{figure*}[!h]
    \centering
    \includegraphics[width=0.99\linewidth]{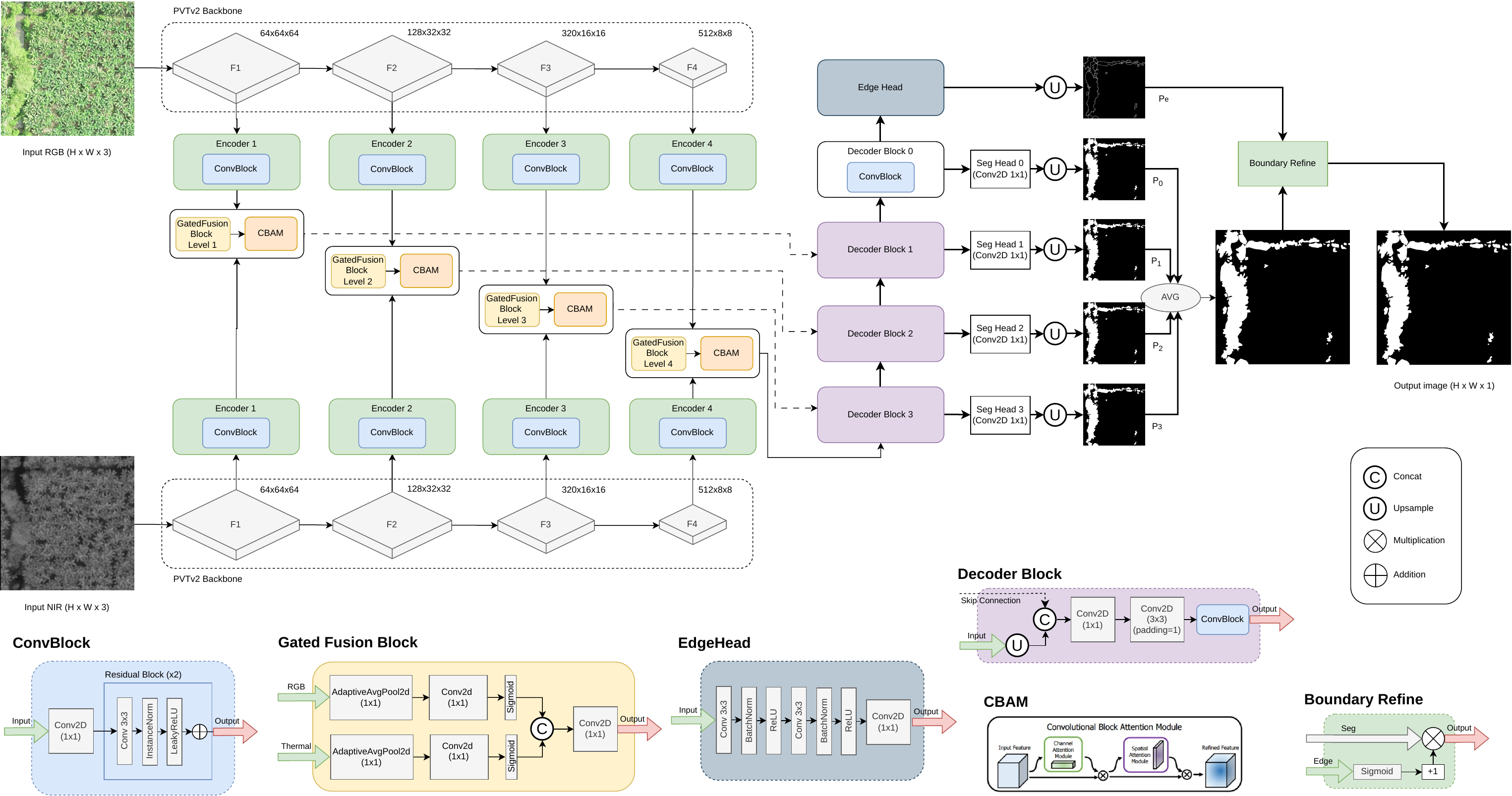}
    \caption{The overall architecture of the proposed SWNet.}
    \label{fig:swnet}
\end{figure*}

\section{Proposed SWNet}
\label{sec:prop}
Similar to most previous works (e.g., \cite{fan2020camouflaged}, \cite{fan2021concealed}, \cite{jiang2022magnet}, \cite{pang2022zoom}, \cite{zhong2022detecting}), the current work adopts an encoder-decoder pipeline to build the proposed SWNet architecture. The proposed architecture, follows a bimodal encoder-decoder structure designed to extract and fuse complementary information from RGB and NIR spectra. SWNet is designed as an end-to-end trainable framework, as illustrated in Fig.~\ref{fig:swnet}.  

\subsection{Feature Extraction Backbone} 
The core of the encoder utilizes the Pyramid Vision Transformer v2~\cite{wang2022pvt} (PVTv2-B2). Unlike traditional CNNs, the PVTv2 leverages a progressive shrinking pyramid and efficient self-attention mechanisms, allowing the model to capture long-range dependencies—a critical factor when identifying camouflaged objects that mimic their immediate surroundings. The backbone extracts features at four different stages, providing a multi-scale representation. % with strides of 4, 8, 16, and 32.

\subsection{Processing Blocks: Residual and ConvBlocks} 
To refine the raw features extracted from the backbone, the architecture employs two fundamental units designed for feature enhancement and dimensional alignment. The ResidualBlock utilizes a $3 \times 3$ convolution followed by Instance Normalization and a LeakyReLU activation, incorporating a skip connection to ensure stable gradient flow and the preservation of fine spatial details. Building upon this, the ConvBlock serves as a bottleneck layer that first projects the backbone’s channel dimensions into the internal feature space via a $1 \times 1$ convolution, subsequently applying a double residual refinement process to further strengthen local feature discrimination and representational power.

\subsection{Convolutional Block Attention Module (CBAM)} 
To suppress background noise and highlight salient regions, the architecture integrates the Convolutional Block Attention Module (CBAM), which processes information through two sequential sub-modules. The process begins with the Channel Gate, which aggregates spatial information by concurrently utilizing Average and Max pooling; these descriptors are then processed by a shared Multi-Layer Perceptron (MLP) to compute a 1D channel attention map, effectively identifying "what" specific features or modalities are most relevant to the task. Subsequently, the Spatial Gate receives these channel-refined features and applies a $7 \times 7$ convolution over the pooled channel axis to generate a 2D spatial mask, which serves to highlight "where" the camouflaged target is likely located within the scene.

\subsection{Bimodal Gated Fusion Module} 
The fusion of RGB and NIR data is executed via a Gated Fusion mechanism that dynamically weights the contribution of each modality to ensure optimal information integration. This process begins with a Gating Mechanism, where independent global average pooling and $1 \times 1$ convolutions are employed to generate modality-specific gates ($\sigma$) that scale the input features based on their relative importance. During the Integration phase, these gated features are concatenated and projected into a unified feature space. Finally, a CBAM pass is applied to the integrated volume, ensuring that the fused representation prioritizes spatial and channel regions where both modalities consistently indicate the presence of a target, thereby significantly reducing the impact of sensor-specific noise or environmental artifacts.

\subsection{Decoder and Feature Aggregation.} The decoder reconstructs the spatial resolution through a series of DecoderBlocks, which progressively restore the feature maps to the original input dimensions. Each block first performs Upsampling via bilinear interpolation to double the spatial resolution, providing a smoother alternative to learnable deconvolution. This is followed by a module, where the upsampled features are concatenated with the fused skip-connections from the encoder stages to recover lost spatial details. Finally, convolutional smoothing is applied to the aggregated volume to refine the feature representation and resolve the "checkerboard" artifacts typically associated with upsampling operations, ensuring a more spatially coherent output for the segmentation heads.

\subsection{Edge-Aware Refinement} 
Detecting the boundaries of camouflaged objects is inherently challenging due to the seamless texture blending between the target and its environment. To address this, the architecture implements an Edge Head branch specifically designed to predict the object's contours from the final decoder stage. These predicted edges are then utilized by the Boundary Refinement module to modulate the primary segmentation mask. Mathematically, the final output is defined as $$O_{final} = Mask \times (1 + \sigma(Edge))$$, where the edge information acts as a regional enhancement factor. This operation enforces sharper transitions at the object boundaries, effectively "carving" the target out of the background and reducing the ambiguity typically found in the peripheral regions of camouflaged targets.

\subsection{Deep Supervision} 
To prevent the vanishing gradient problem and encourage feature learning at multiple scales, Deep Supervision is applied. Four auxiliary segmentation heads produce intermediate masks from different decoder stages. During inference, these masks are averaged to produce a robust, multi-scale prediction, which is then refined by the edge map to yield the final output.

\subsection{Loss Function} 
Following the methodology of previous studies \cite{fan2021concealed,sun2021c2fnet}, this work adopts the loss function introduced by \cite{wei2020f3net}. The SWNet decoder generates a set of predictions denoted as $\{P_i\}_{i=0}^{3}$. During training, each prediction $P_i$ is upsampled to match the original input dimensions and supervised using a combination of Binary Cross-Entropy ($\mathcal{L}_{BCE}$) \cite{de2005tutorial} and Intersection over Union ($\mathcal{L}_{IoU}$) \cite{mattyus2017deeproadmapper} losses. Consistent with \cite{fan2021concealed}, the total objective is calculated by aggregating losses across multiple stages. Furthermore, an Edge Loss is incorporated to refine the boundary detection branch, employing a binary cross-entropy objective against an edge ground truth ($GT_{edge}$) derived from the mask $GT$ via morphological operations (specifically, the difference between local max and min pooling). The total loss function for SWNet is defined as:

\begin{equation}
\mathcal{L}(P,GT) = \sum_{i=0}^{3} \mathcal{L}_{BCE}(P_i,GT) + \mathcal{L}_{IoU}(P_i,GT)
\end{equation}

\begin{equation}
\mathcal{L}_{total} = \mathcal{L}(P,GT) + \mathcal{L}_{edge}(E, GT_{edge})
\end{equation}

\subsection{Implementation Details}
The proposed SWNet is implemented using the PyTorch framework. It employs the PVTv2-B2 backbone \cite{wang2022pvt}, pretrained on ImageNet, as the primary encoder. Network optimization is performed using the AdamW optimizer with a weight decay of $1\times10^{-4}$. The learning rate is initialized at $1\times10^{-4}$ and adjusted following a cosine annealing schedule. All input images are resized to $416 \times 416$ for both the training and inference phases. The model is trained end-to-end for 200 epochs with a batch size of 10 on a GeForce RTX 4090 (24 GB) GPU. The complete training code is publicly available on GitHub.

\subsection{Datasets}
\begin{figure*}[!h]
\setlength\tabcolsep{0.75pt}
\centering
\scalebox{1.0}{
%\resizebox{\columnwidth}{!}{
\begin{tabular}{ccccc}

\includegraphics[width=.16\textwidth, height=2.4cm]{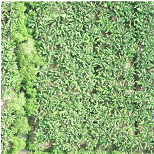} & 
\includegraphics[width=.16\textwidth, height=2.4cm]{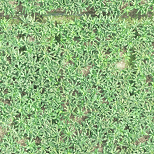} &
\includegraphics[width=.16\textwidth, height=2.4cm]{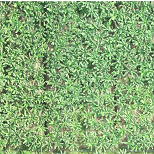} &
\includegraphics[width=.16\textwidth, height=2.4cm]{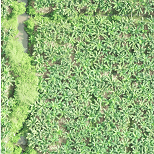} &
\includegraphics[width=.16\textwidth, height=2.4cm]{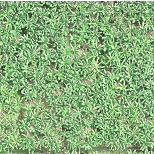} \\

\includegraphics[width=.16\textwidth, height=2.4cm]{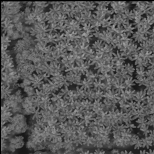} & 
\includegraphics[width=.16\textwidth, height=2.4cm]{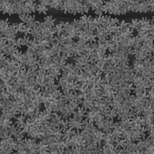} &
\includegraphics[width=.16\textwidth, height=2.4cm]{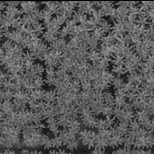} &
\includegraphics[width=.16\textwidth, height=2.4cm]{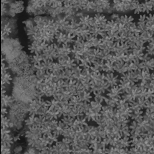} &
\includegraphics[width=.16\textwidth, height=2.4cm]{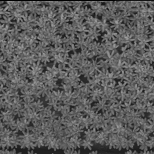} \\

\includegraphics[width=.16\textwidth, height=2.4cm]{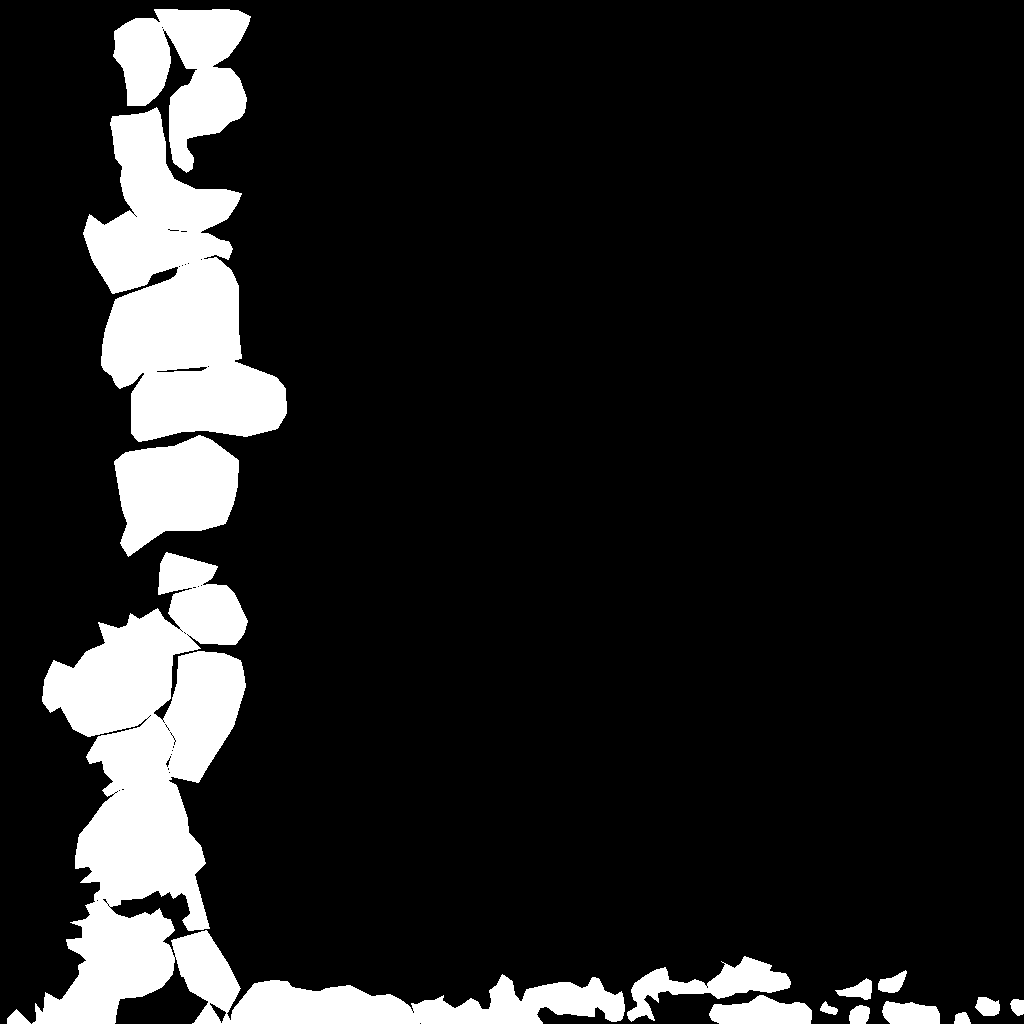} & 
\includegraphics[width=.16\textwidth, height=2.4cm]{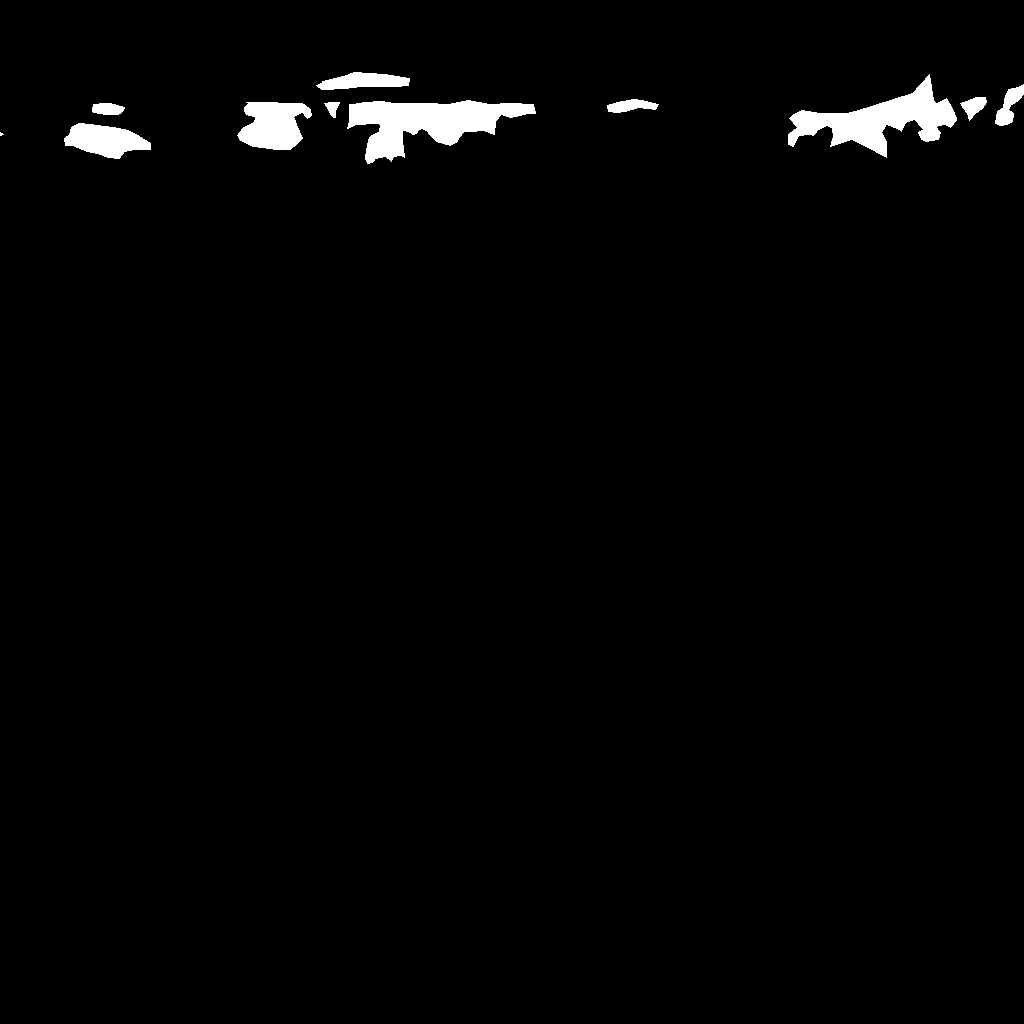} &
\includegraphics[width=.16\textwidth, height=2.4cm]{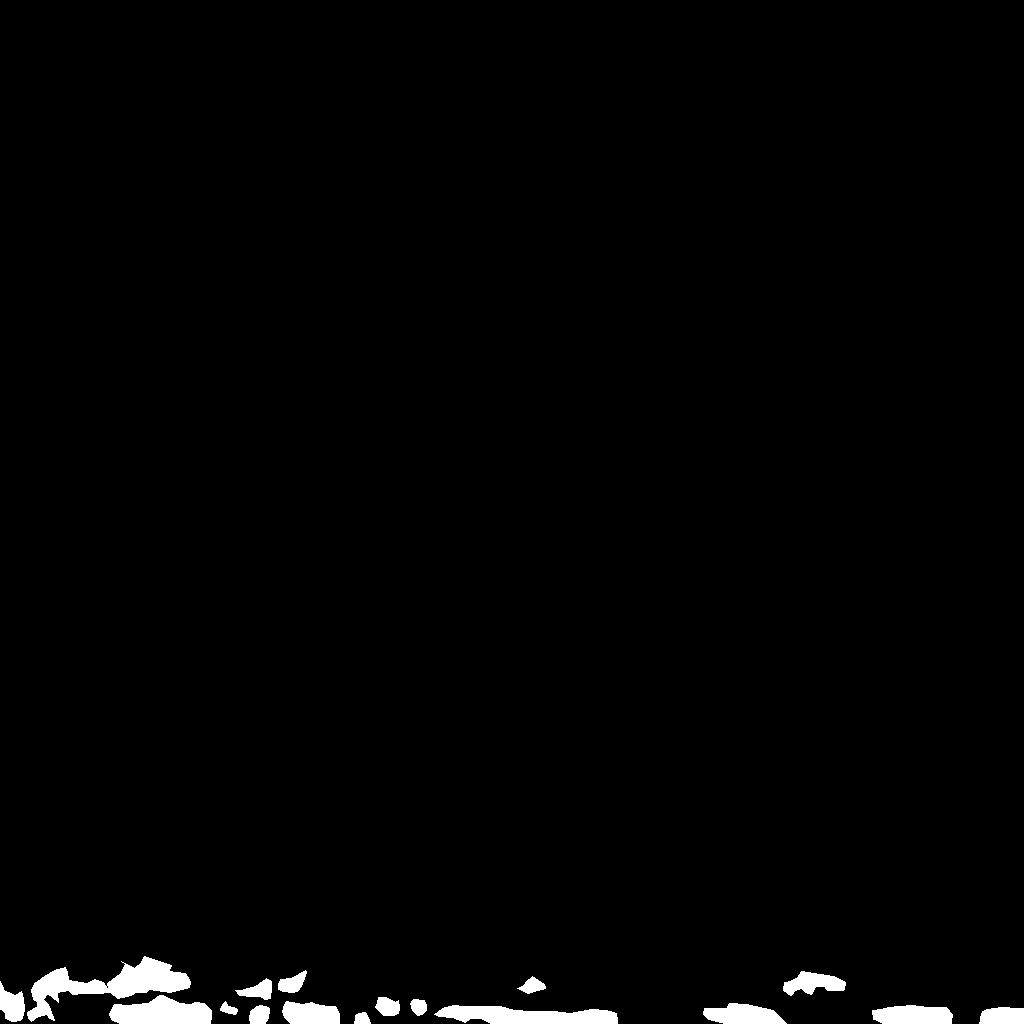} &
\includegraphics[width=.16\textwidth, height=2.4cm]{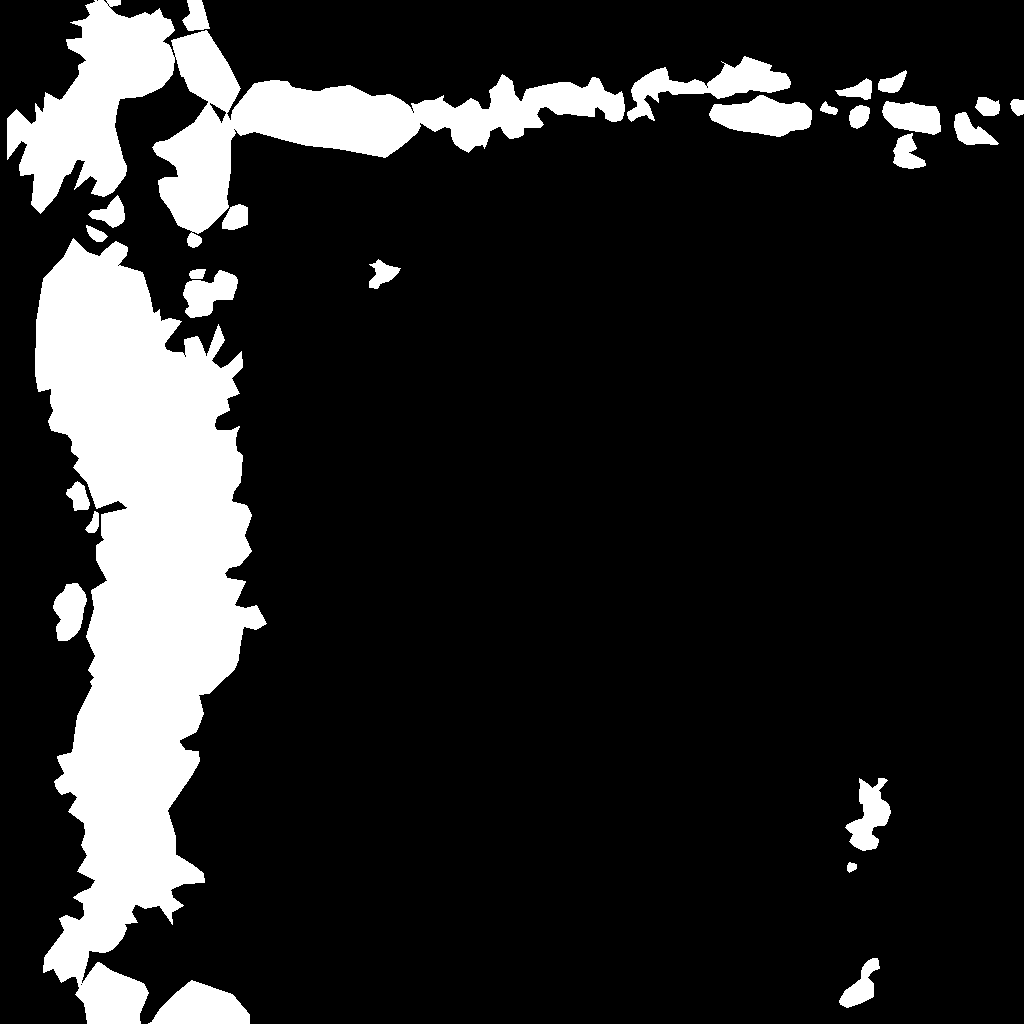} &
\includegraphics[width=.16\textwidth, height=2.4cm]{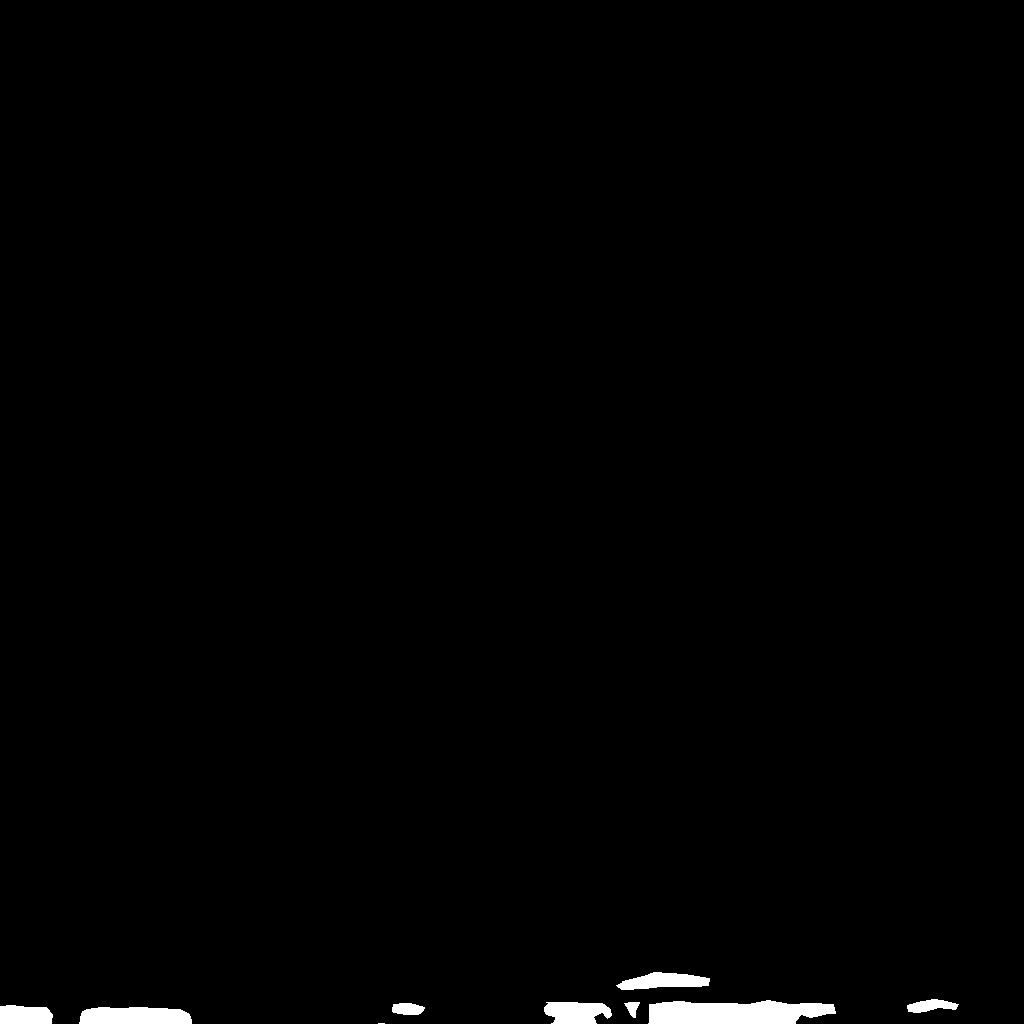} \\

\end{tabular}
}
%}
\caption{Example of RGB, NIR and mask images of the Weeds-Banana \cite{velesaca2026unveiling} dataset.}
\label{fig:rgb_nir_images}
\end{figure*}

To validate the proposed architecture, a state-of-the-art multispectral dataset, Weeds-Banana \cite{velesaca2026unveiling} is used. The dataset consists of 272 high-resolution image pairs ($1024 \times 1024$ pixels) providing a specialized benchmark for camouflaged weed detection in dense agricultural canopies. Figure \ref{fig:rgb_nir_images} shows examples of RGB, NIR, and mask images of the Weeds-Banana dataset\footnote{https://www.kaggle.com/datasets/hvelesaca/weedbananacod}.

\subsection{Evaluation Metrics}
\label{subSec:metricsEval}
To provide a multidimensional assessment of camouflaged object detection (COD) performance, five standard evaluation metrics are employed. The structural similarity between the predicted maps and the ground truth (GT) is quantified using the Structure-measure ($S_\alpha$), while the weighted F-measure ($F^w_\beta$) offers an enhanced evaluation by incorporating spatial weights that emphasize boundary accuracy. Pixel-level discrepancies are measured through the Mean Absolute Error ($M$), calculated between the normalized prediction and the GT. Furthermore, the E-measure ($E_\phi$) is utilized to capture both global and local accuracy based on human visual perception mechanisms, alongside the F-measure ($F_\beta$), which serves as the harmonic mean of precision and recall. To ensure a robust performance comparison across different models, the mean values for the latter two metrics ($E^{mean}_\phi$ and $F^{mean}_\beta$) are reported, as calculated across various thresholds.In order to guarantee a fair comparison with existing state-of-the-art methods, the training configurations for all evaluated models have been standardized. Comprehensive details regarding the specific optimizers, learning rates, schedulers, and loss functions utilized for each network within the benchmark are provided in Table~\ref{tab:networks_params}.

\begin{table*}[!h]
    \centering
    %\scriptsize
    \caption{Metric evaluation results for each COD technique on the Weeds-Banana \cite{velesaca2026unveiling} dataset, reported for the RGB and NIR baseline. Results are presented using the metric notation defined in Sec.~\ref{subSec:metricsEval}, ``$\uparrow/\downarrow$'' indicates that larger or smaller is better. The best three performing results are highlighted using color: \First{First}, \Second{Second}, and \Third{Third} respectively.}
    \resizebox{2\columnwidth}{!}{
    \begin{tabular}{l|l|rrrrrrrrr}
        \toprule
        Technique & Input & $S_\alpha \uparrow$ & $F^{w}_\beta \uparrow$ & $M \downarrow$ & $E^{adp}_\phi \uparrow$ & $E^{mean}_\phi \uparrow$ & $E^{max}_\phi \uparrow$ & $F^{adp}_\beta \uparrow$ & $F^{mean}_\beta \uparrow$ & $F^{max}_\beta \uparrow$ \\
        
        \midrule
        %\multirow{2}{*}{BASNet \cite{qin2019basnet}} & Vis & 0.9228 & 0.8696 & 0.0072 & 0.9694 & 0.9755 & 0.9931 & 0.8185 & 0.8724 & 0.9006 \\
        %& NIR & - & - & - & - & - & - & - & - & - \\
        
        \multirow{2}{*}{SINet-V2 \cite{fan2021concealed}} & Vis & 0.7917 & 0.5754 & 0.0179 & 0.8052 & 0.9175 & 0.9709 & 0.5382 & 0.6153 & 0.6753 \\
        & NIR & 0.7522 & 0.5272 & 0.0215 & 0.7668 & 0.8773 & 0.9322 & 0.4945 & 0.5646 & 0.6347 \\
        
        \midrule
        
        \multirow{2}{*}{BGNet \cite{chen2022boundary}} & Vis & 0.7600 & 0.4362 & 0.0319 & \third{0.9543} & 0.9468 & \first{0.9946} & \third{0.8022} & \third{0.8469} & \first{0.9298} \\
        & NIR & 0.7444 & 0.4217 & 0.0341 & 0.9017 & 0.9340 & \third{0.9910} & 0.7221 & 0.7962 & \second{0.8950} \\
        
        \midrule
        
        \multirow{2}{*}{C$^{2}$F-Net \cite{chen2022camouflaged}} & Vis & 0.6107 & 0.2293 & 0.0827 & 0.7488 & 0.8065 & 0.9729 & 0.5231 & 0.6072 & 0.7848 \\
        & NIR & 0.6277 & 0.2522 & 0.0710 & 0.7731 & 0.8274 & 0.9662 & 0.5438 & 0.6188 & 0.7577 \\

        \midrule
        
        \multirow{2}{*}{OCENet \cite{liu2022modeling}} & Vis & 0.8207 & 0.7185 & 0.0133 & 0.9455 & 0.9416 & 0.9791 & 0.7151 & 0.7643 & 0.7835 \\
        & NIR & 0.7651 & 0.5876 & 0.0175 & 0.9124 & 0.9341 & 0.9814 & 0.6070 & 0.6380 & 0.6575 \\

        \midrule
        
        \multirow{2}{*}{EAMNet \cite{sun2023edge}} & Vis & 0.5796 & 0.1953 & 0.1031 & 0.7151 & 0.7685 & 0.9122 & 0.4630 & 0.4895 & 0.5943 \\
        & NIR & 0.5467 & 0.1488 & 0.1232 & 0.7324 & 0.7701 & 0.9308 & 0.4393 & 0.4215 & 0.5255 \\

        \midrule
        
        \multirow{2}{*}{DGNet \cite{ji2023deep}} & Vis & 0.8439 & 0.6908 & 0.0142 & 0.8299 & 0.9302 & 0.9768 & 0.6015 & 0.7120 & 0.7787 \\ 
        & NIR & 0.8381 & 0.6829 & 0.0151 & 0.8530 & 0.9433 & 0.9855 & 0.6119 & 0.7117 & 0.7631 \\
        
        \midrule

        \multirow{2}{*}{HitNet \cite{hu2023high}} & Vis & 0.8773 & 0.8090 & \third{0.0088} & 0.9291 & 0.9652 & \second{0.9937} & 0.7393 & 0.7970 & 0.8582 \\
        & NIR & 0.8705 & 0.7992 & 0.0089 & 0.9455 & 0.9463 & 0.9546 & 0.7813 & 0.8045 & 0.8264 \\
        
        %\multirow{2}{*}{PCNet \cite{yang2024plantcamo}} & Vis & 0.9257 & 0.8933 & 0.0054 & 0.9723 & 0.9862 & 0.9950 & 0.8360 & 0.8851 & 0.9194 \\
        %& NIR & - & - & - & - & - & - & - & - & - \\
        %\cmidrule(lr){2-12}
        \midrule
        
        \multirow{2}{*}{ARNet \cite{wang2025assisted}} & Vis & 0.8800 & \third{0.8131} & 0.0091 & \second{0.9827} & 0.9604 & 0.9904 & \second{0.8153} & \second{0.8492} & 0.8653 \\
        & NIR & 0.8265 & 0.7229 & 0.0127 & 0.9429 & 0.9402 & 0.9860 & 0.7138 & 0.7670 & 0.7920 \\
        
        \midrule
        
        \multirow{2}{*}{CHNet \cite{wang2025efficient}} & Vis & \third{0.8839} & 0.8058 & 0.0090 & 0.9291 & \second{0.9713} & 0.9873 & 0.7358 & 0.8096 & 0.8619 \\
        & NIR & 0.8027 & 0.6971 & 0.0124 & 0.9267 & 0.9112 & 0.9838 & 0.7200 & 0.7423 & 0.7631 \\

        \midrule
         
        \multirow{2}{*}{ARNet-v2 \cite{wang2025assistedv2}} & Vis & \first{0.9027} & \second{0.8229} & \second{0.0086} & 0.9493 & \third{0.9667} & 0.9901 & 0.7681 & 0.8425 & 0.8737 \\
        & NIR & 0.8027 & 0.6971 & 0.0124 & 0.9267 & 0.9112 & 0.9838 & 0.7200 & 0.7423 & 0.7631 \\

        \midrule
        
        \multirow{3}{*}{SWNet (ours)} & Vis & 0.7971 & 0.7227 & 0.0122 & 0.9305 & 0.9338 & 0.9391 & 0.7040 & 0.7175 & 0.7419 \\ 
        & NIR & 0.8413 & 0.7624 & 0.0108 & 0.9300 & 0.9325 & 0.9385 & 0.7332 & 0.7460 & 0.7676 \\
        & Vis+NIR & \second{0.8966} & \first{0.8767} & \first{0.0070} & \first{0.9857} & \first{0.9860} & 0.9906 & \first{0.8493} & \first{0.8590} & \third{0.8788} \\

        \bottomrule

    \end{tabular}
    }
    \label{tab:results_cod_weedsbanana}
\end{table*}

\begin{figure*}[!h]
\setlength\tabcolsep{0.75pt}
\centering
\scalebox{1.0}{
\begin{tabular}{ccccccccc}

\rotatebox{90}{\scriptsize{RGB}} & 
\includegraphics[width=.16\textwidth, height=2.21cm]{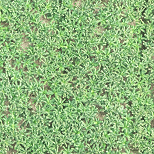} & 
\includegraphics[width=.16\textwidth, height=2.21cm]{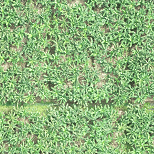} & 
\includegraphics[width=.16\textwidth, height=2.21cm]{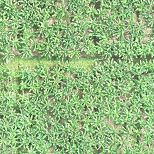} & 
\includegraphics[width=.16\textwidth, height=2.21cm]{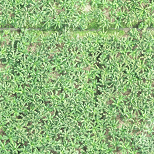} &  
\includegraphics[width=.16\textwidth, height=2.21cm]{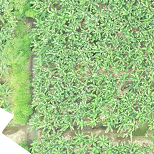} &  
\includegraphics[width=.16\textwidth, height=2.21cm]{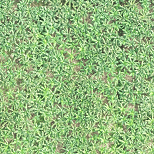} \\ 

\rotatebox{90}{\scriptsize{NIR}} & 
\includegraphics[width=.16\textwidth, height=2.21cm]{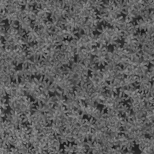} & 
\includegraphics[width=.16\textwidth, height=2.21cm]{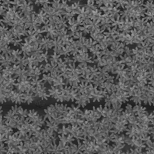} & 
\includegraphics[width=.16\textwidth, height=2.21cm]{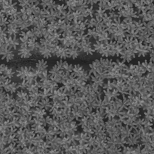} & 
\includegraphics[width=.16\textwidth, height=2.21cm]{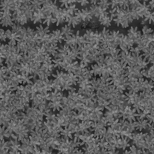} &  
\includegraphics[width=.16\textwidth, height=2.21cm]{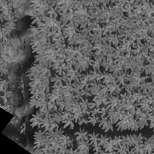} &  
\includegraphics[width=.16\textwidth, height=2.21cm]{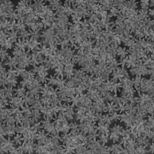} \\  

\rotatebox{90}{\scriptsize{GT}} & 
\includegraphics[width=.16\textwidth, height=2.21cm]{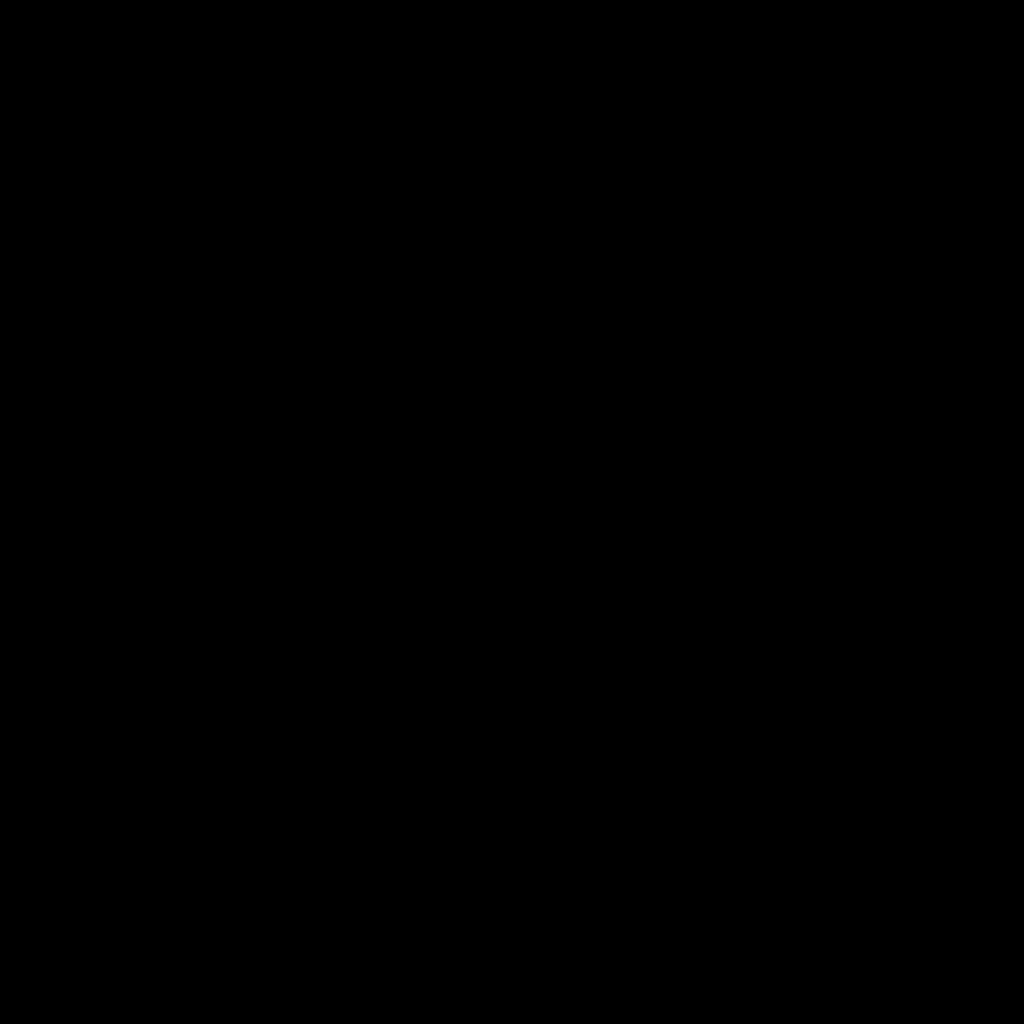} & 
\includegraphics[width=.16\textwidth, height=2.21cm]{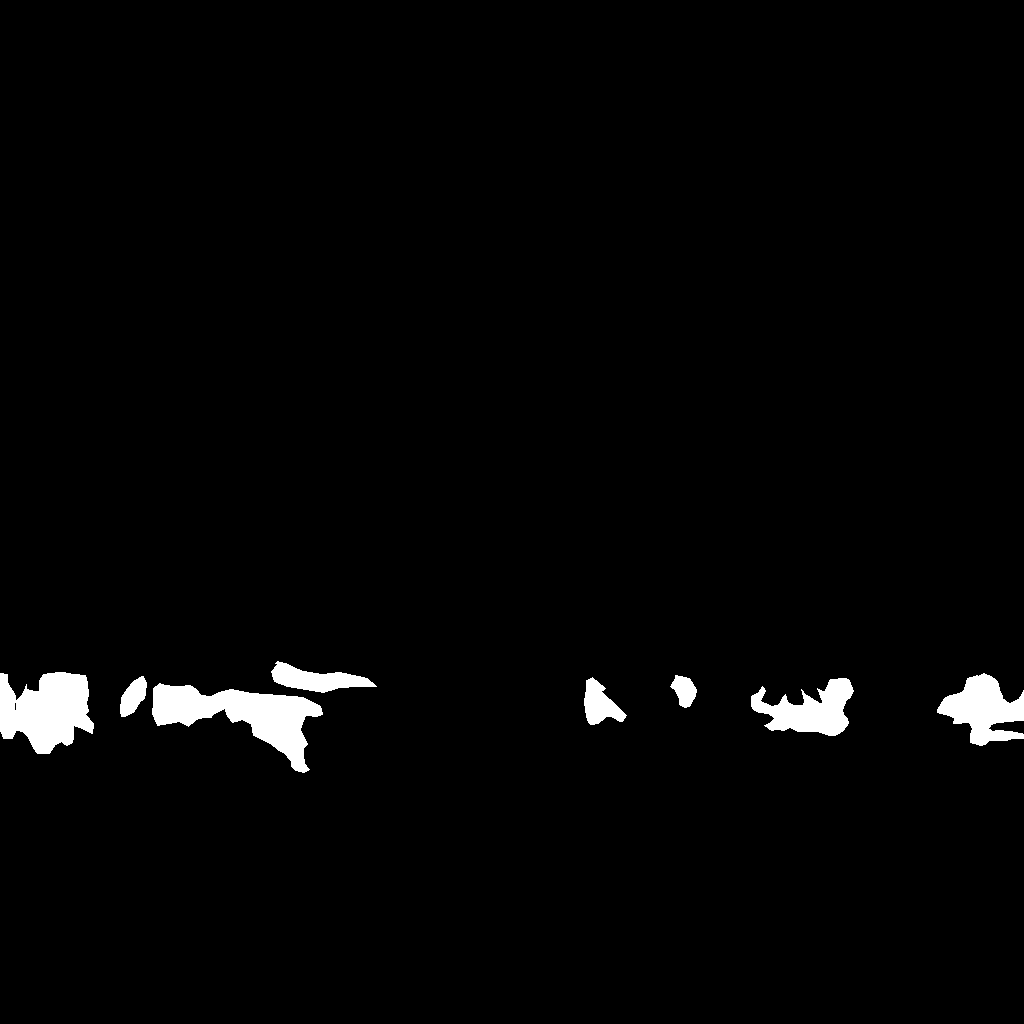} & 
\includegraphics[width=.16\textwidth, height=2.21cm]{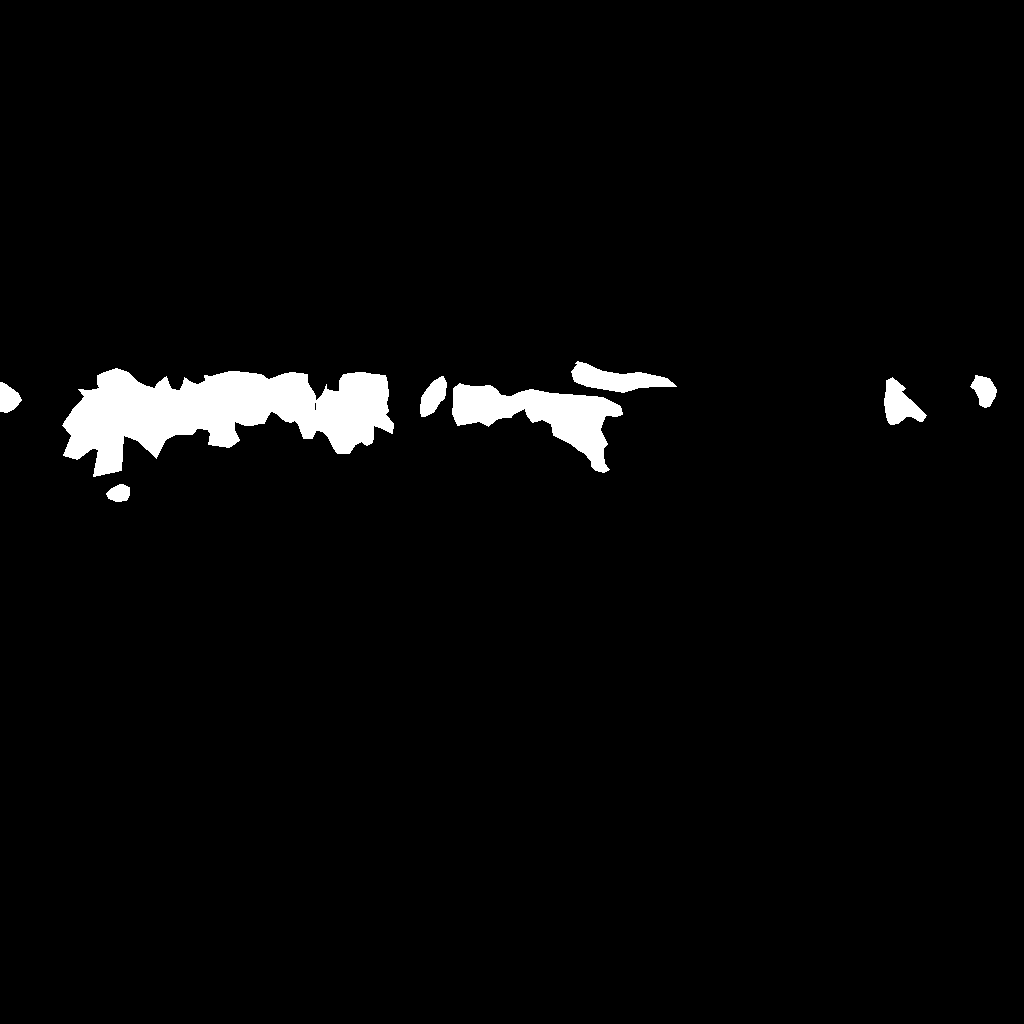} & 
\includegraphics[width=.16\textwidth, height=2.21cm]{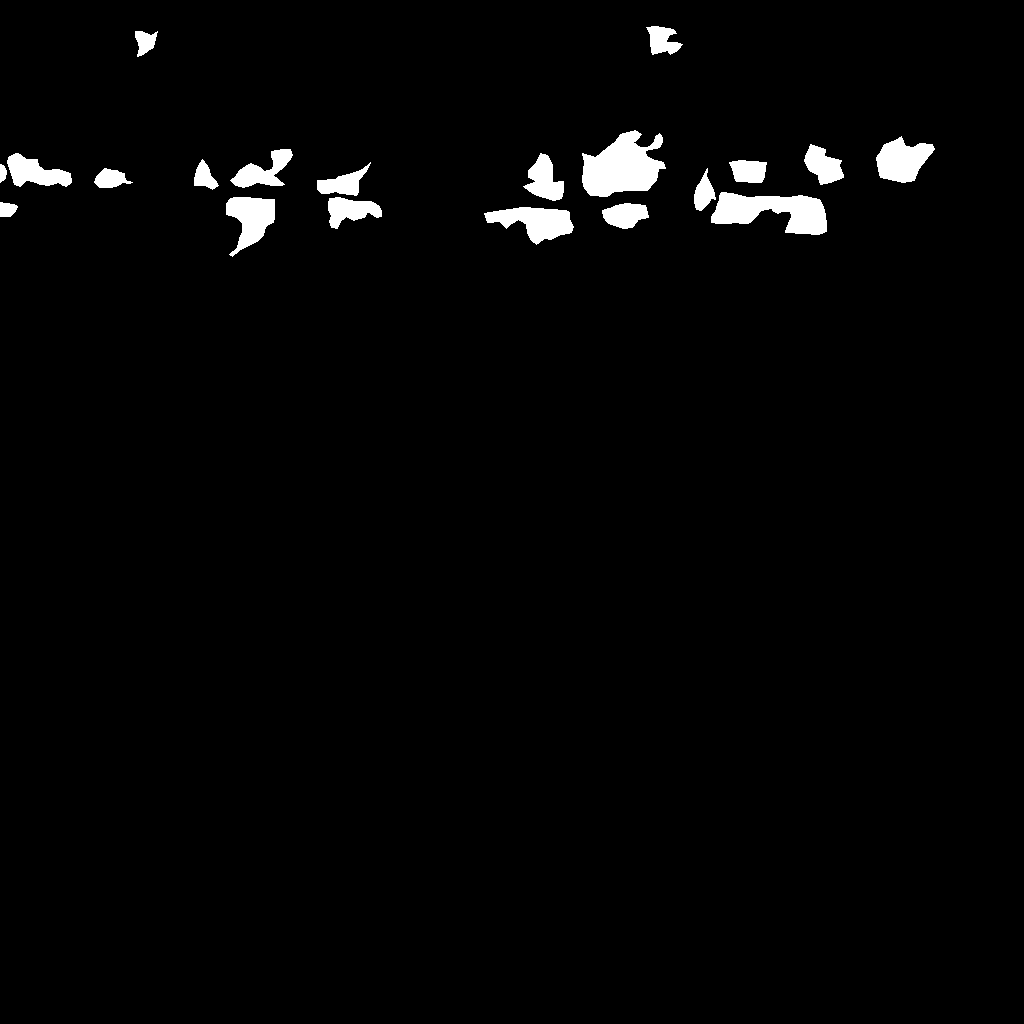} &  
\includegraphics[width=.16\textwidth, height=2.21cm]{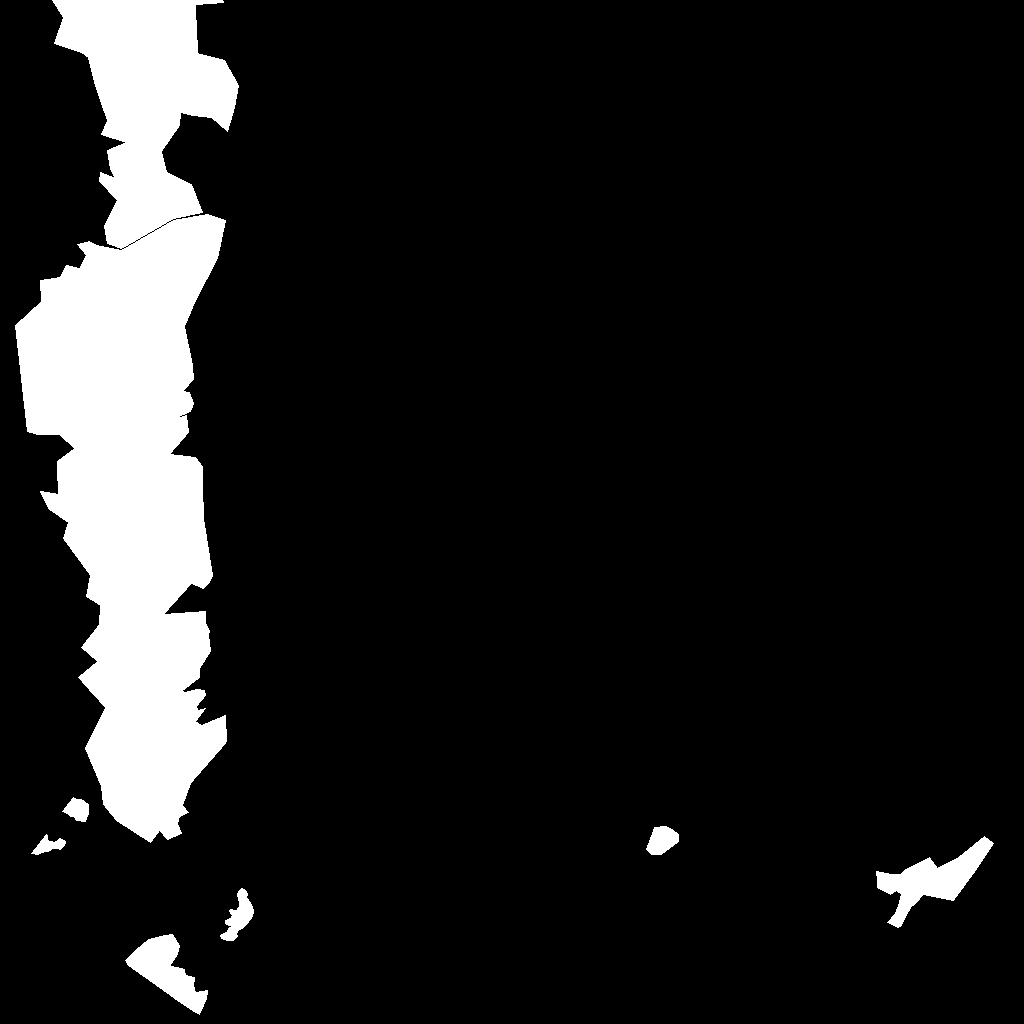} &  
\includegraphics[width=.16\textwidth, height=2.21cm]{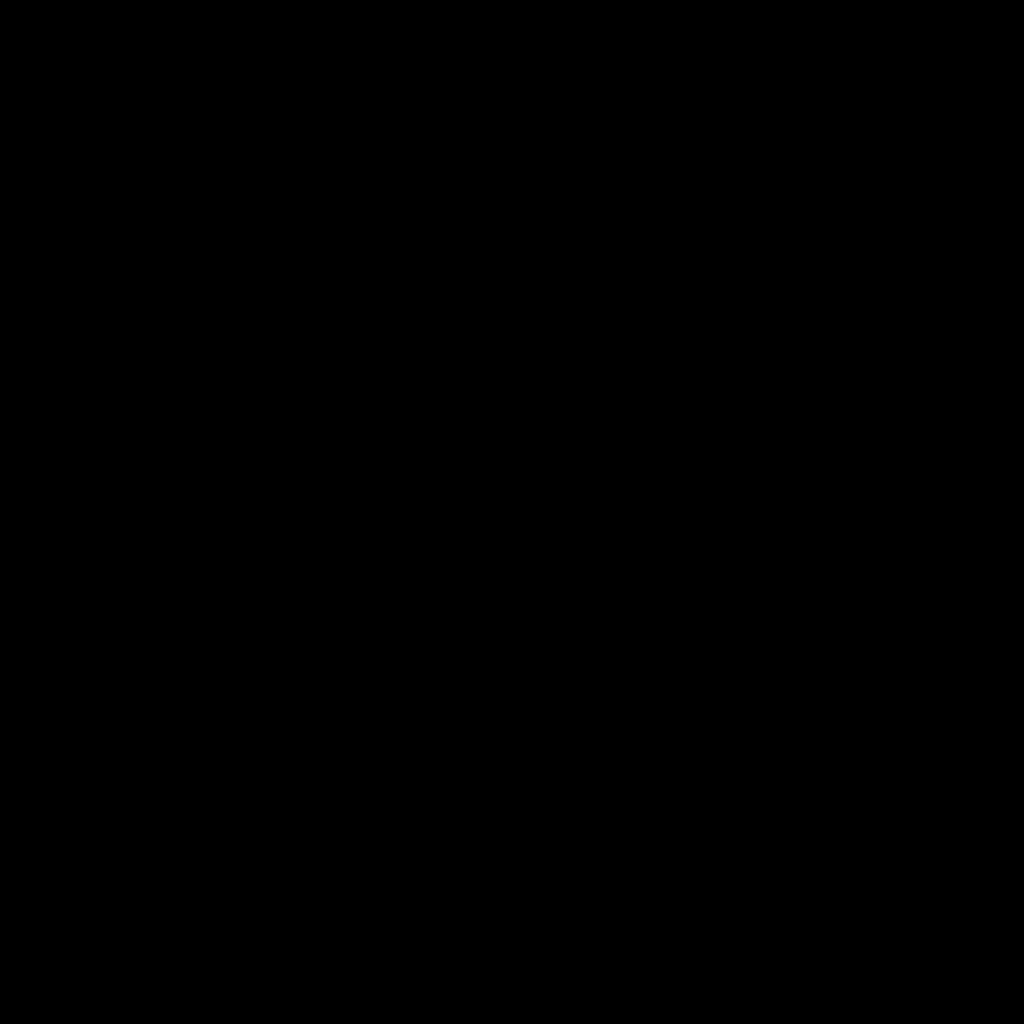} \\ 

\rotatebox{90}{\scriptsize{BGNet Vis \cite{chen2022boundary}}} & 
\includegraphics[width=.16\textwidth, height=2.21cm]{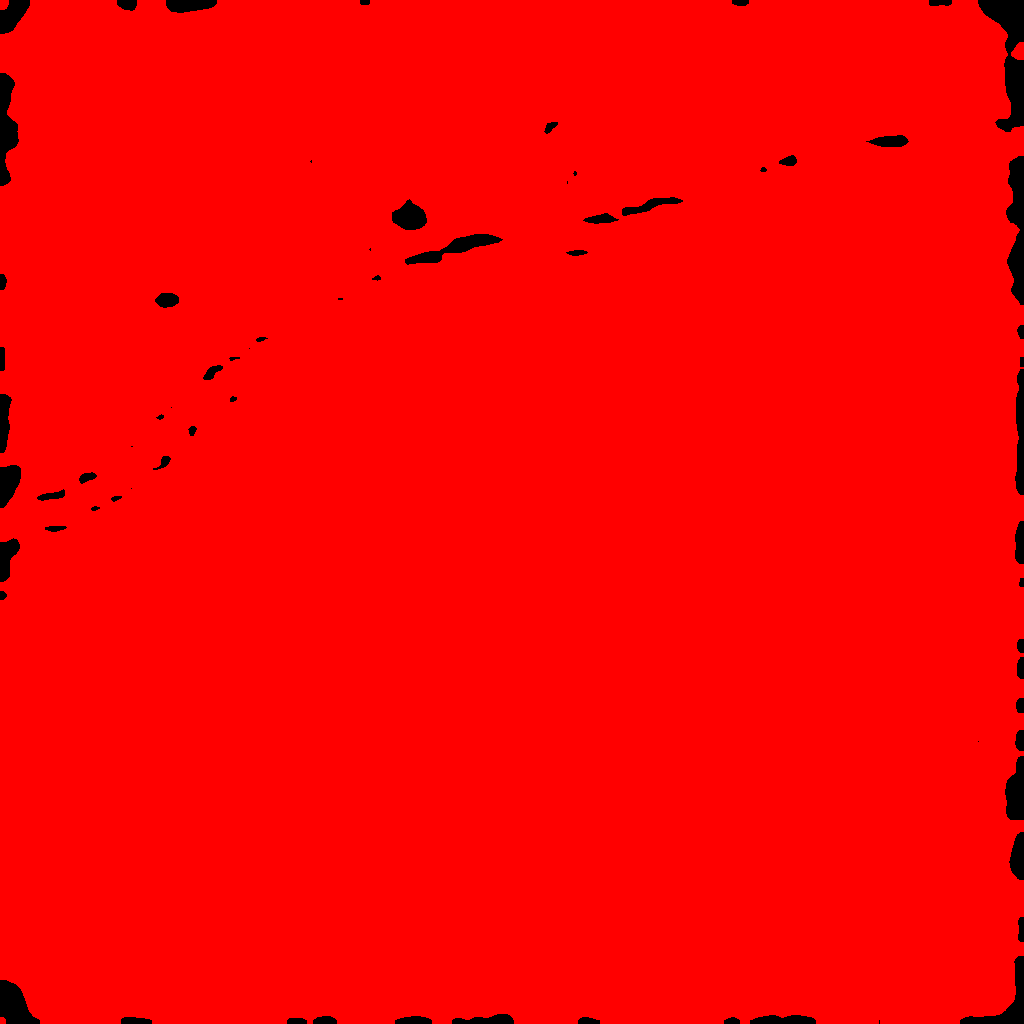} & 
\includegraphics[width=.16\textwidth, height=2.21cm]{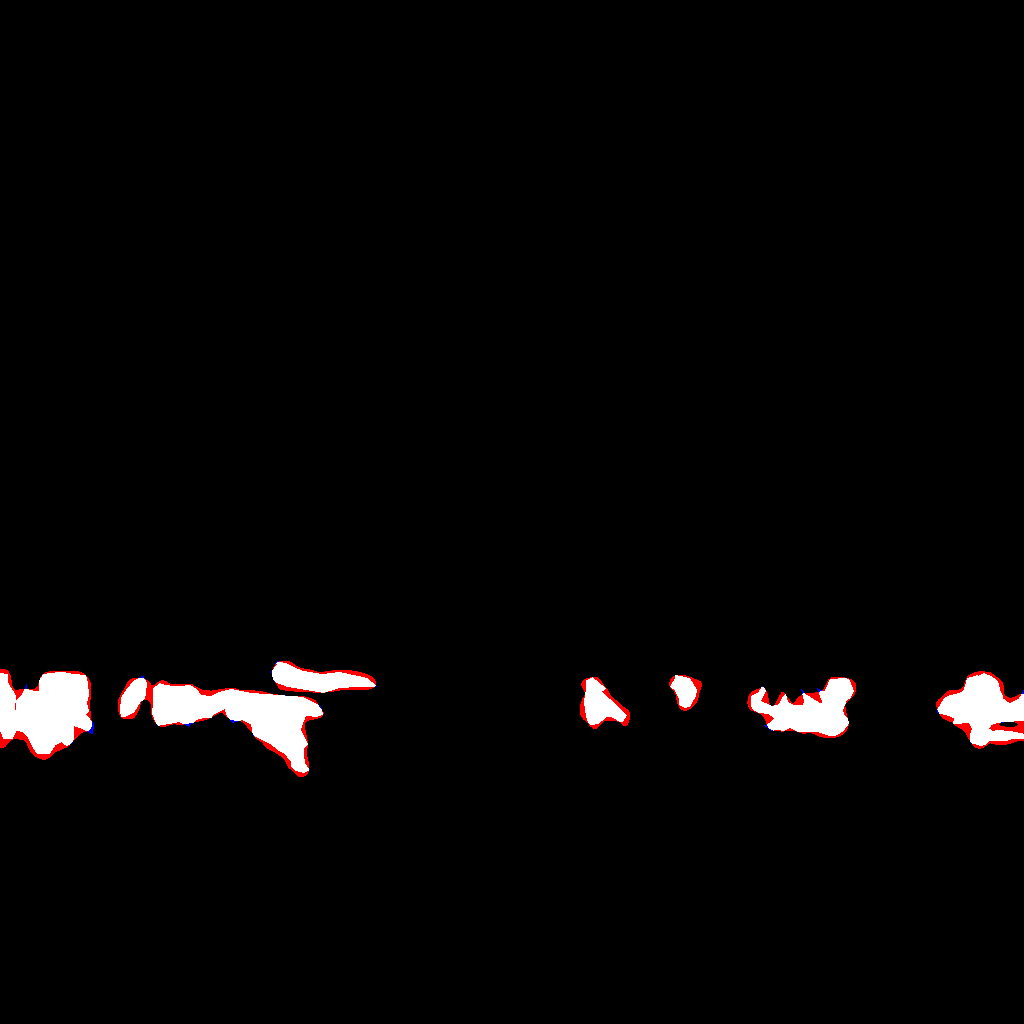} & 
\includegraphics[width=.16\textwidth, height=2.21cm]{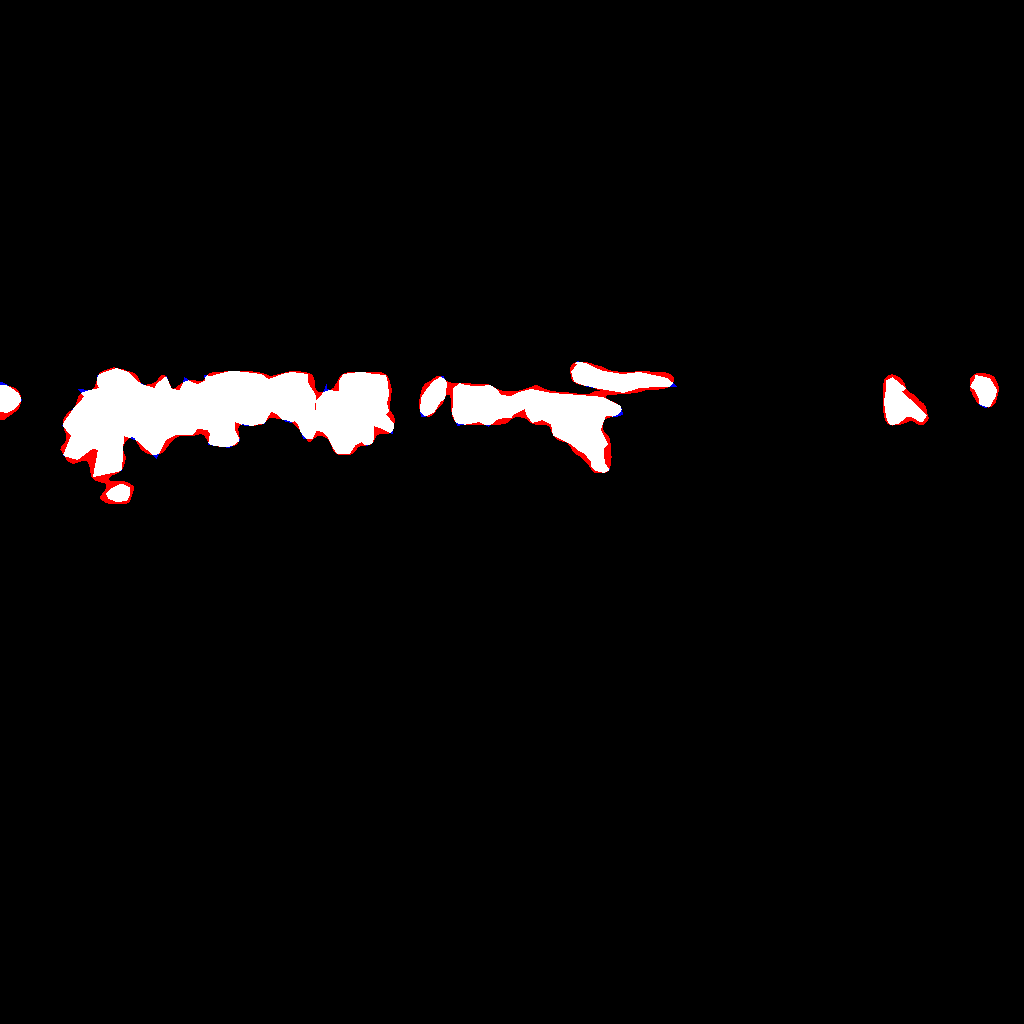} & 
\includegraphics[width=.16\textwidth, height=2.21cm]{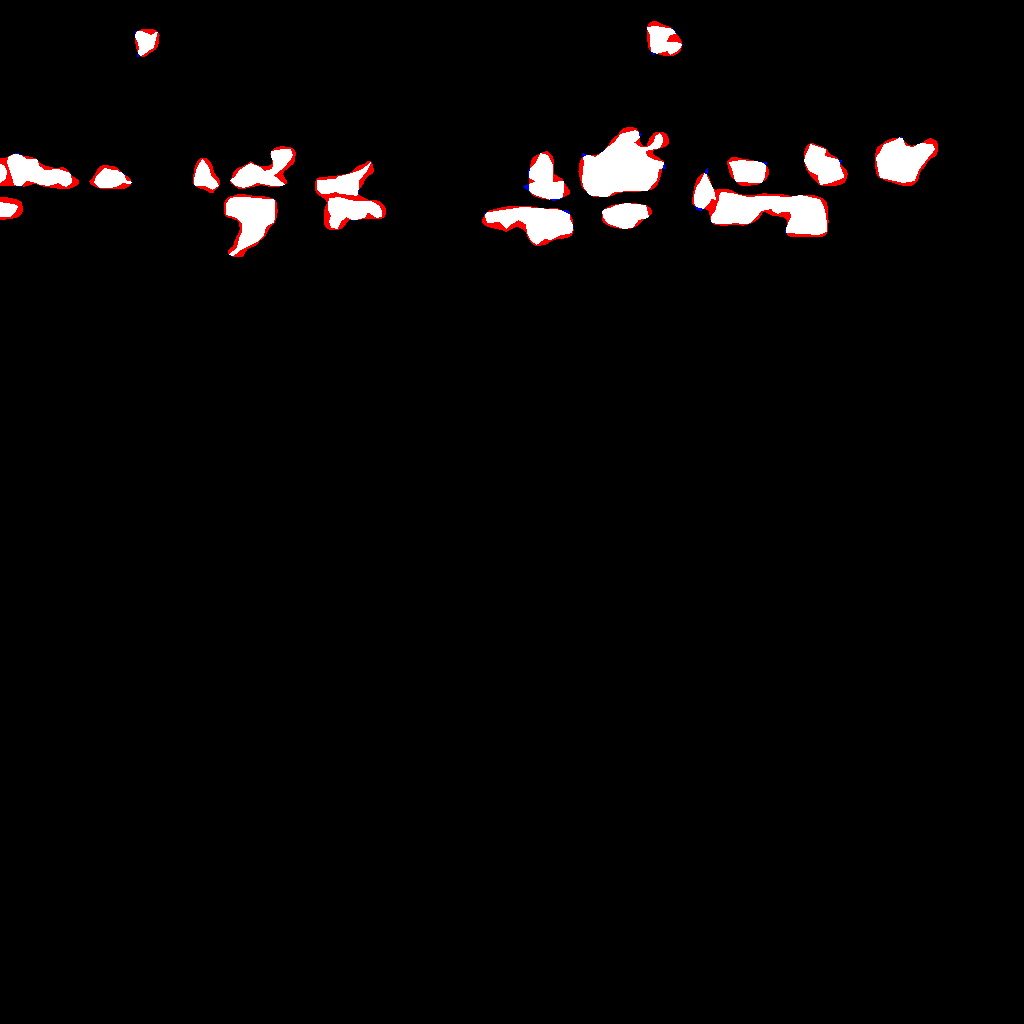} & 
\includegraphics[width=.16\textwidth, height=2.21cm]{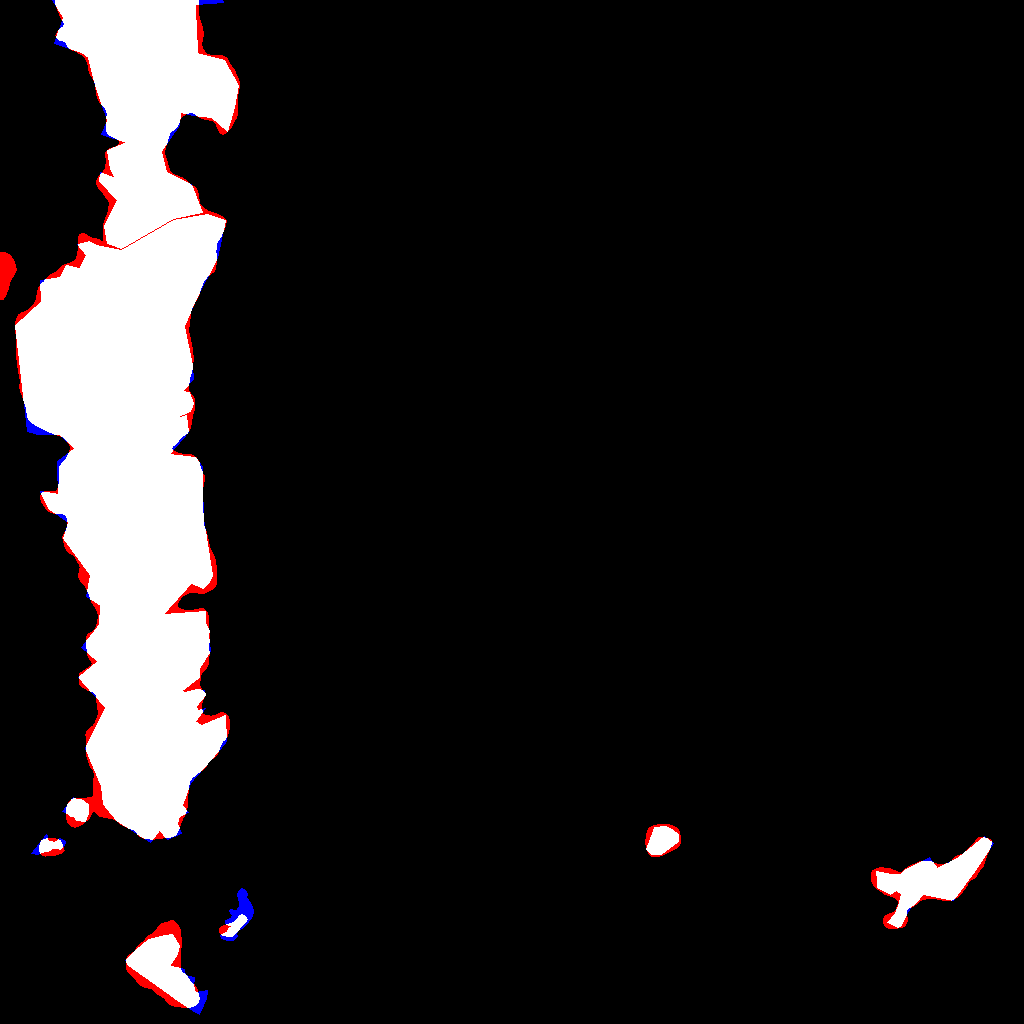} &  
\includegraphics[width=.16\textwidth, height=2.21cm]{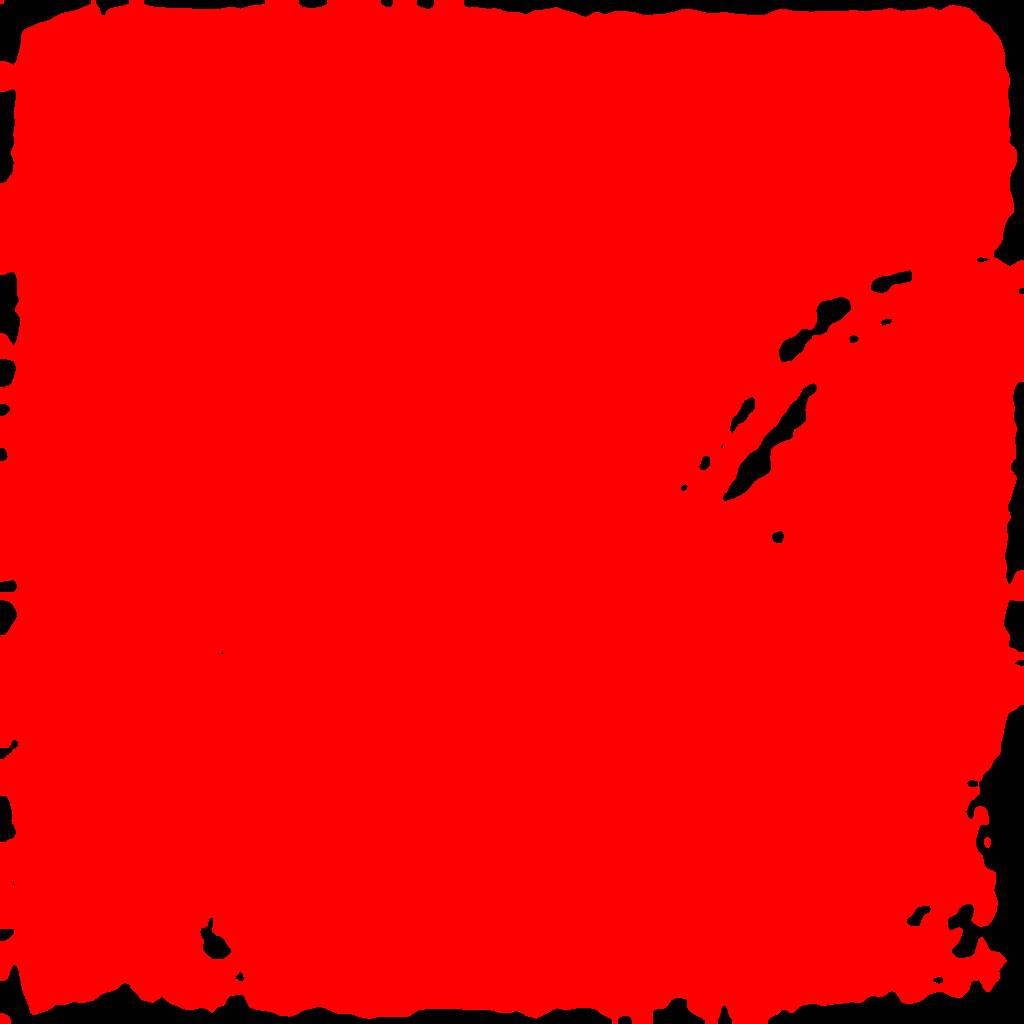} \\ 

\rotatebox{90}{\scriptsize{HitNet Vis \cite{hu2023high}}} & 
\includegraphics[width=.16\textwidth, height=2.21cm]{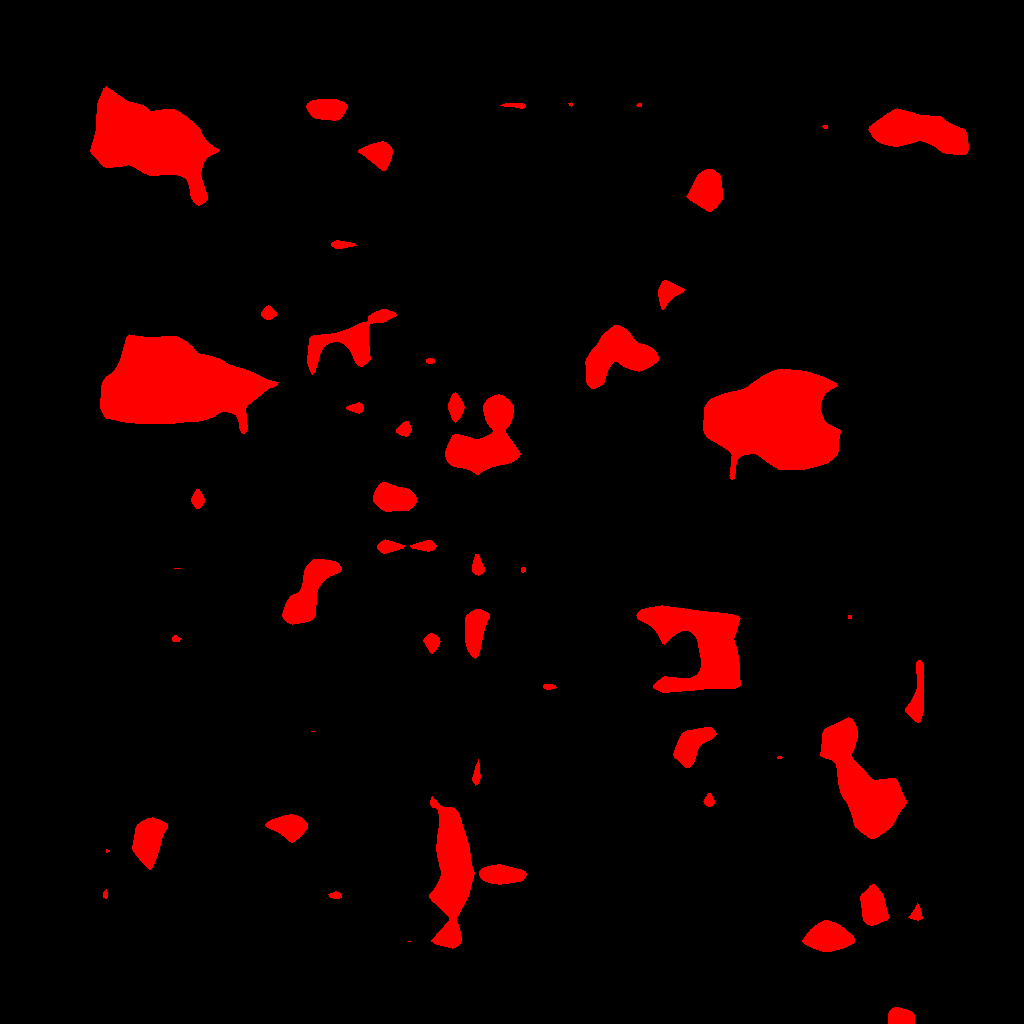} & 
\includegraphics[width=.16\textwidth, height=2.21cm]{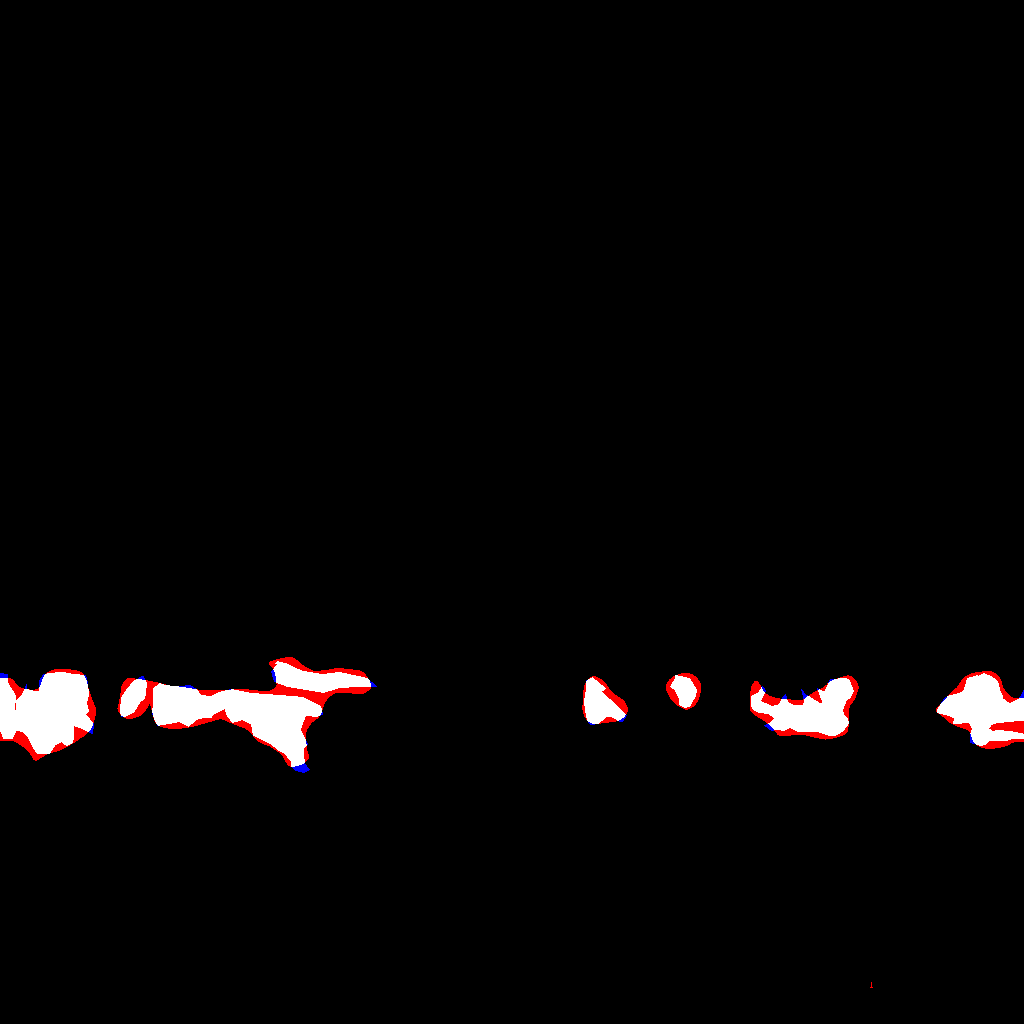} & 
\includegraphics[width=.16\textwidth, height=2.21cm]{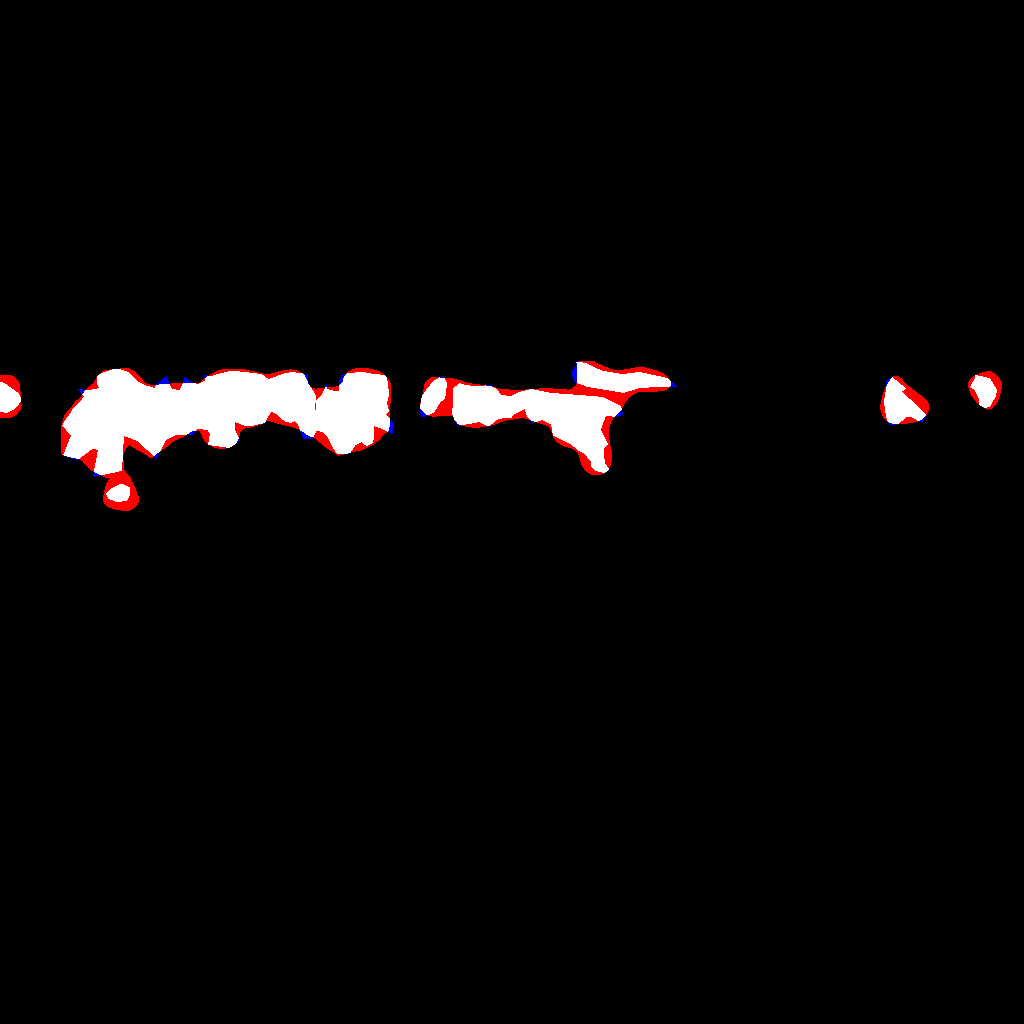} & 
\includegraphics[width=.16\textwidth, height=2.21cm]{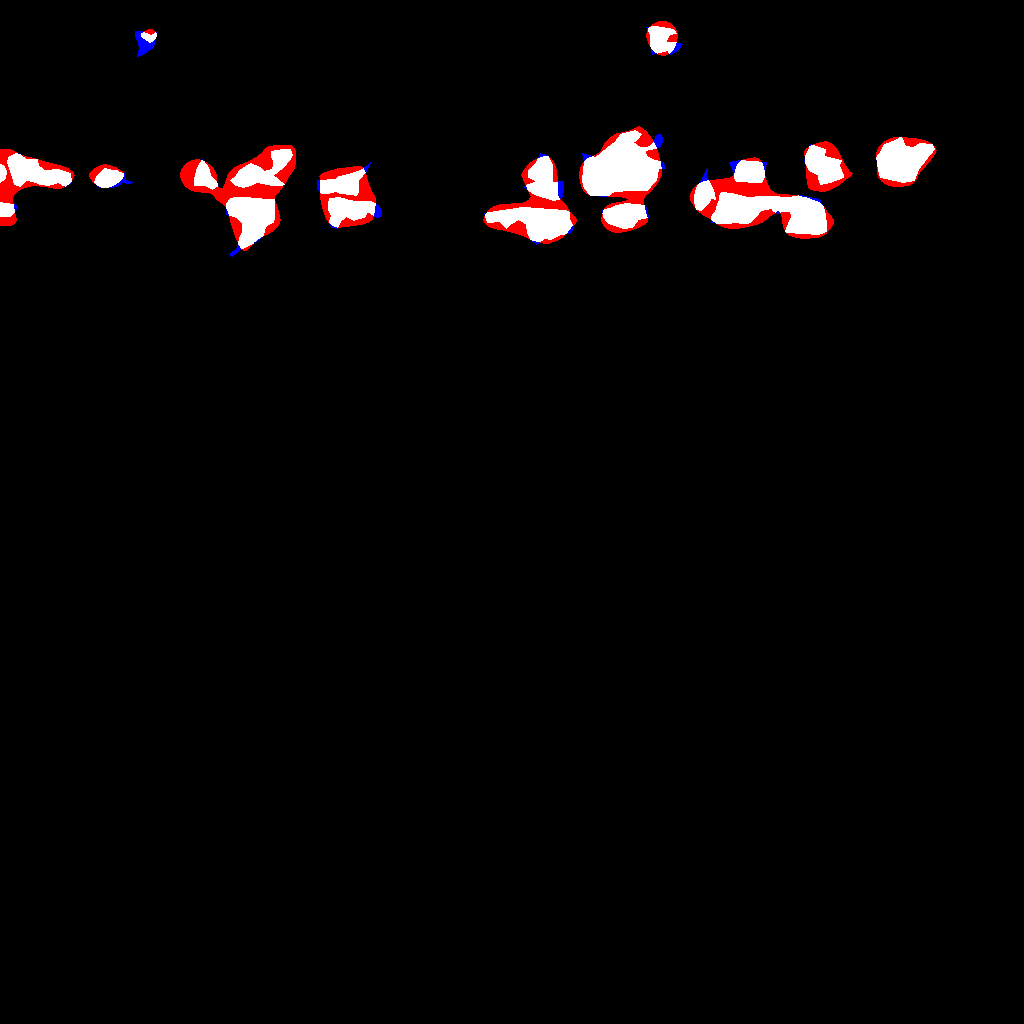} & 
\includegraphics[width=.16\textwidth, height=2.21cm]{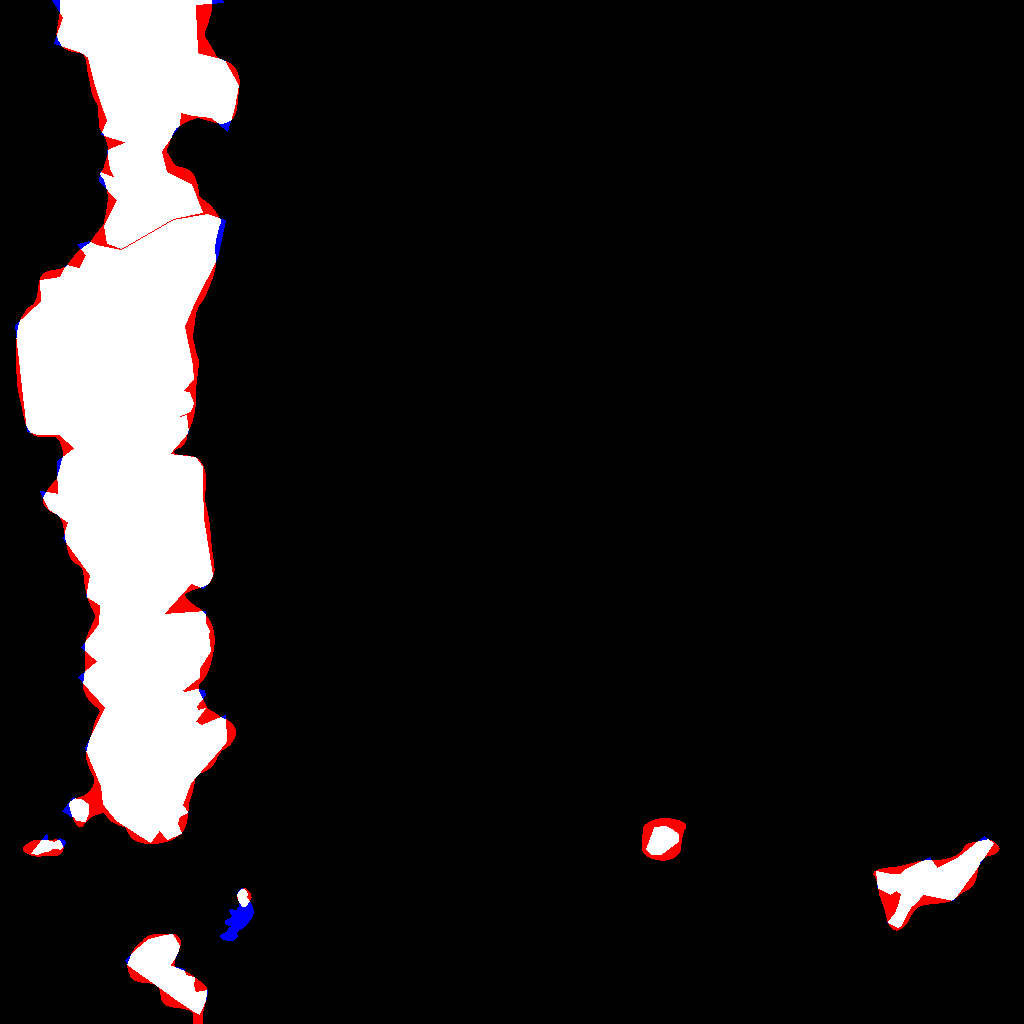} &  
\includegraphics[width=.16\textwidth, height=2.21cm]{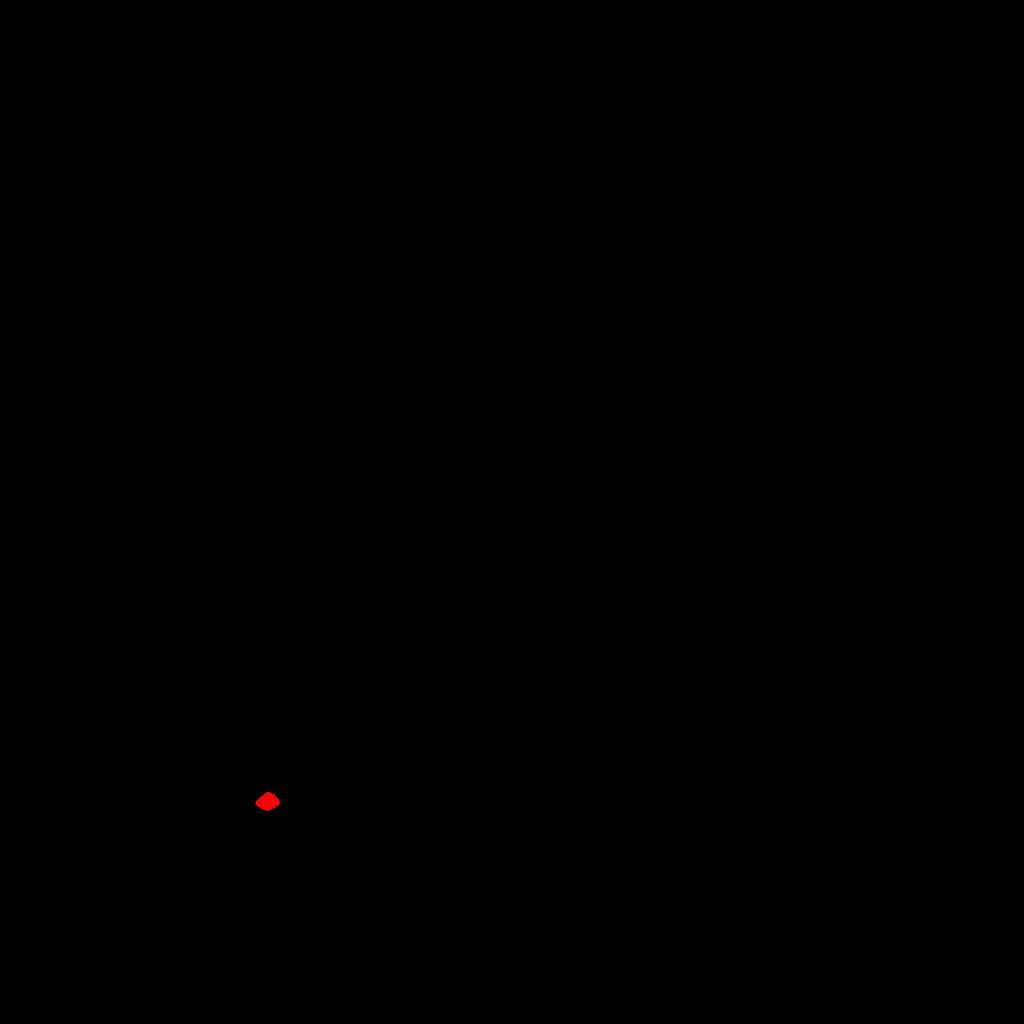} \\ 

\rotatebox{90}{\scriptsize{ARNet Vis \cite{wang2025assisted}}} & 
\includegraphics[width=.16\textwidth, height=2.21cm]{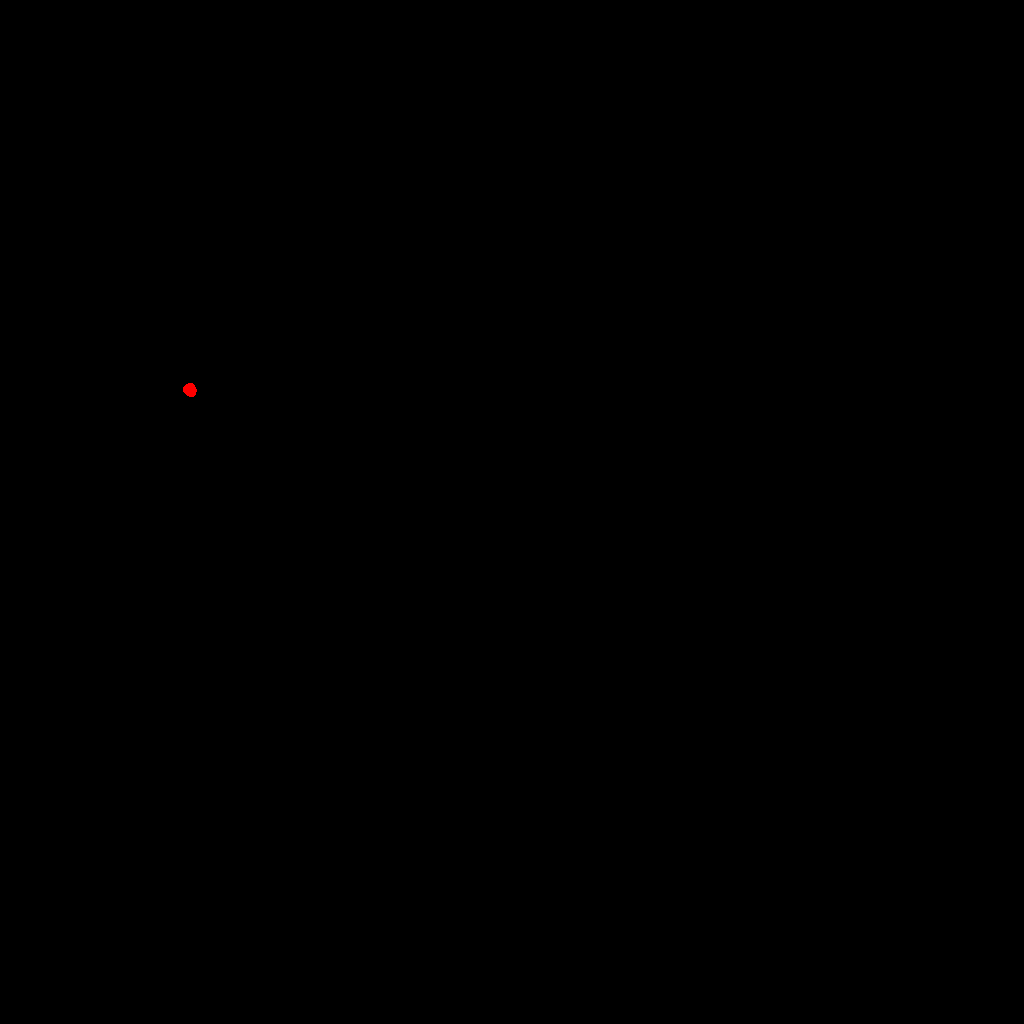} & 
\includegraphics[width=.16\textwidth, height=2.21cm]{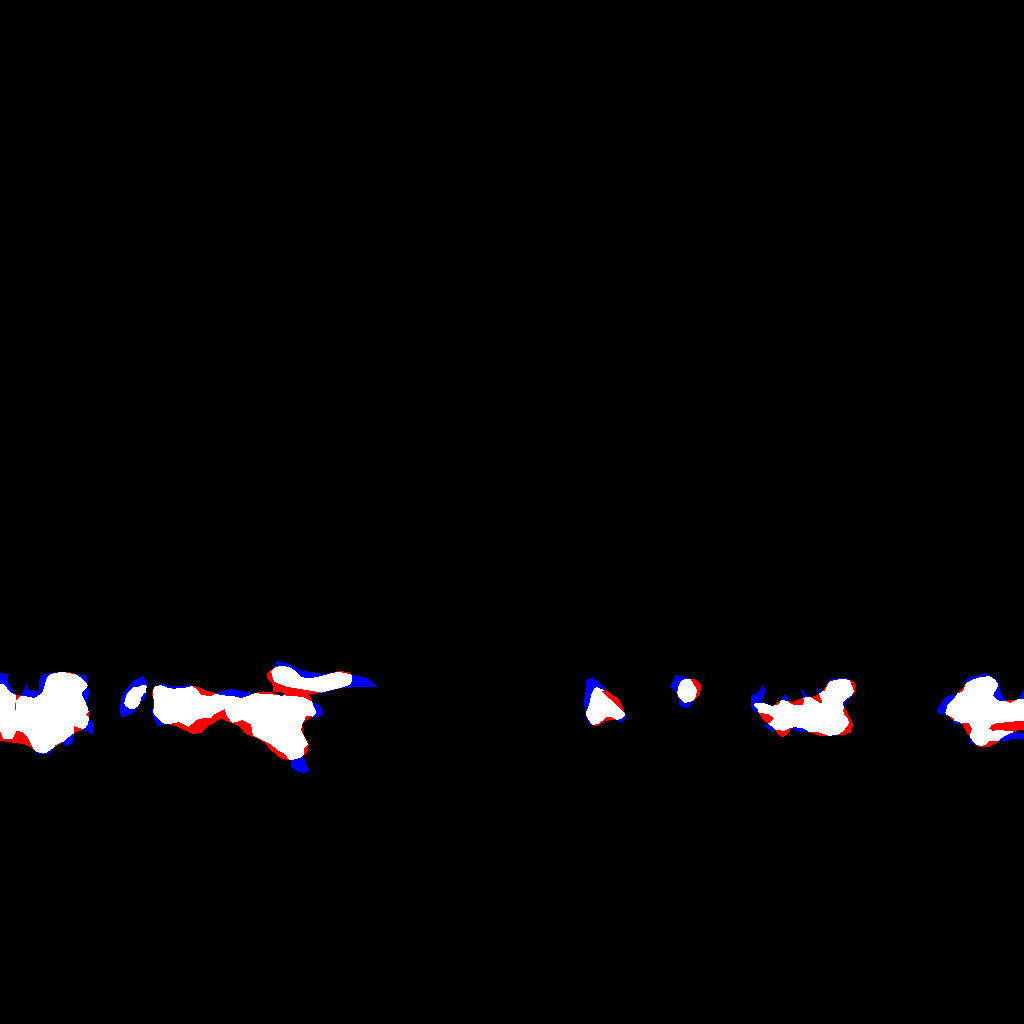} & 
\includegraphics[width=.16\textwidth, height=2.21cm]{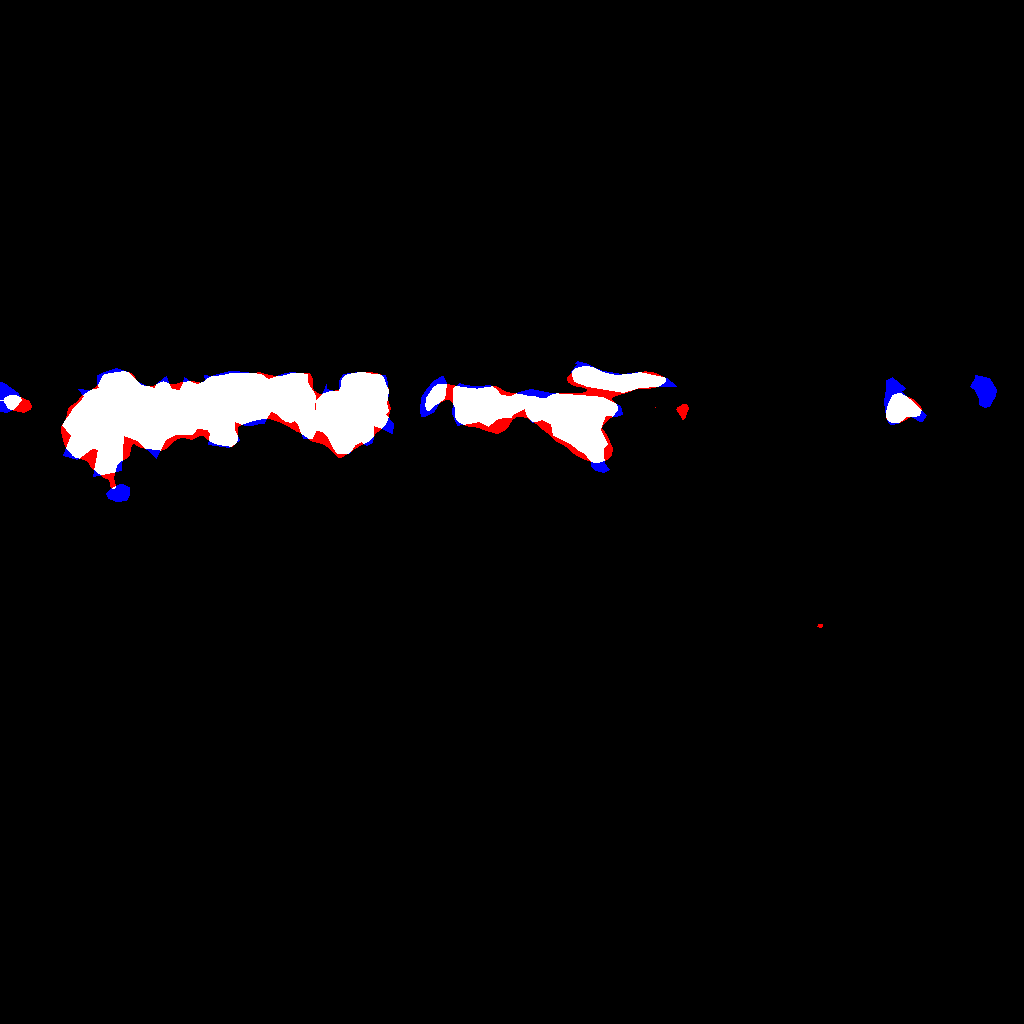} & 
\includegraphics[width=.16\textwidth, height=2.21cm]{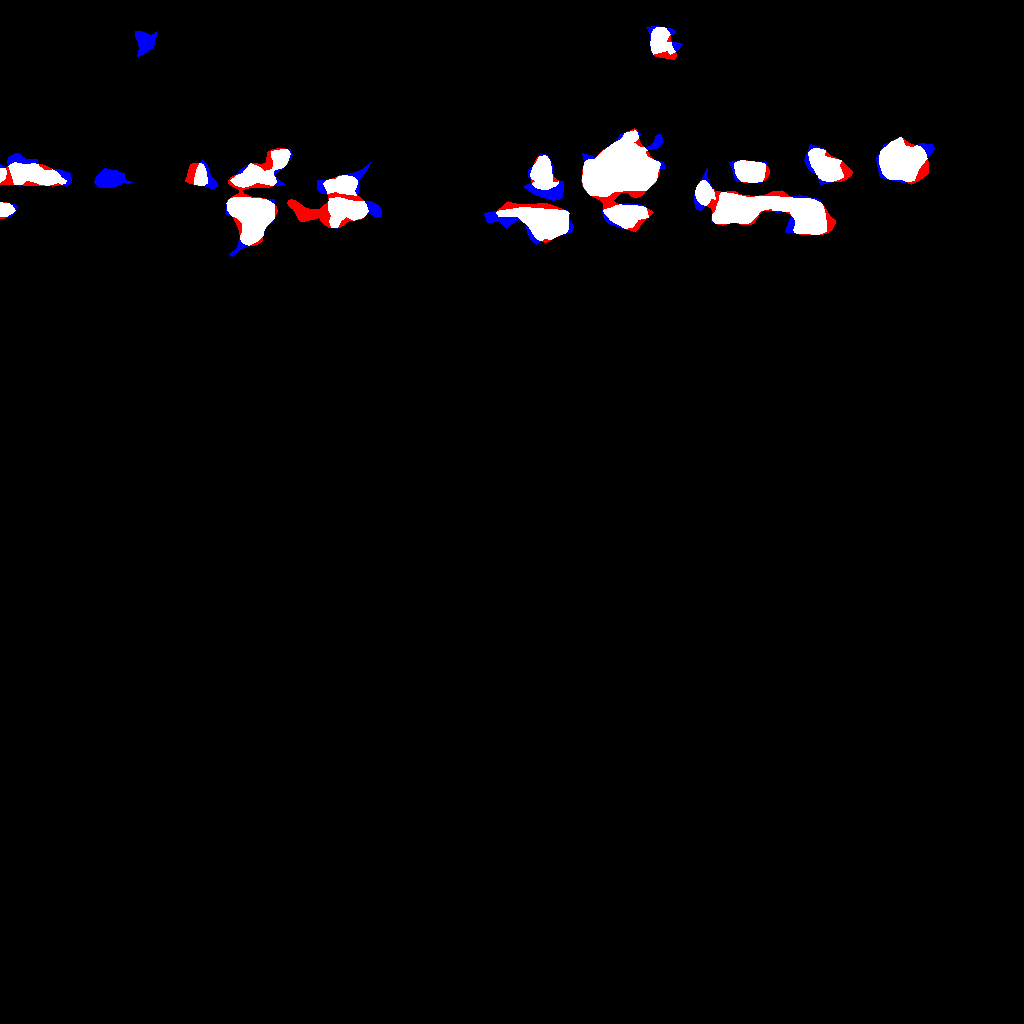} & 
\includegraphics[width=.16\textwidth, height=2.21cm]{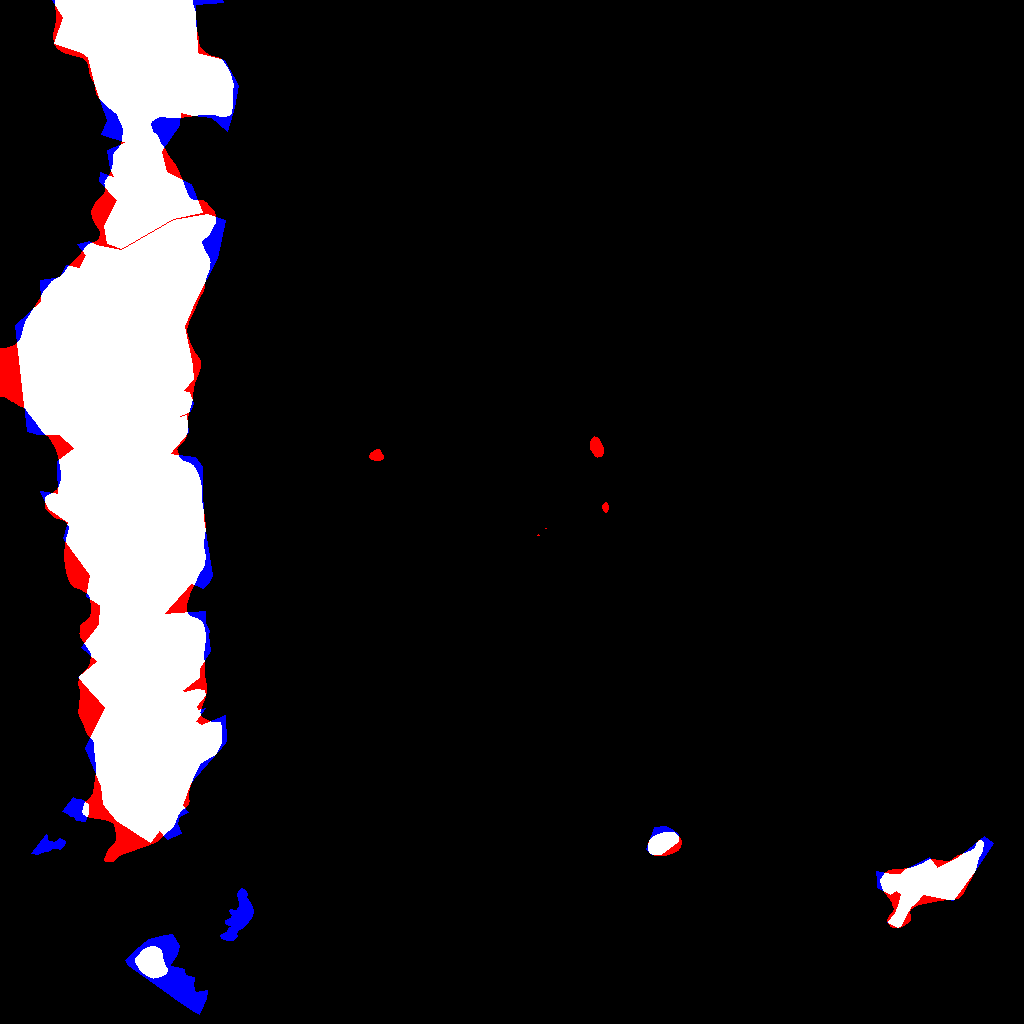} &  
\includegraphics[width=.16\textwidth, height=2.21cm]{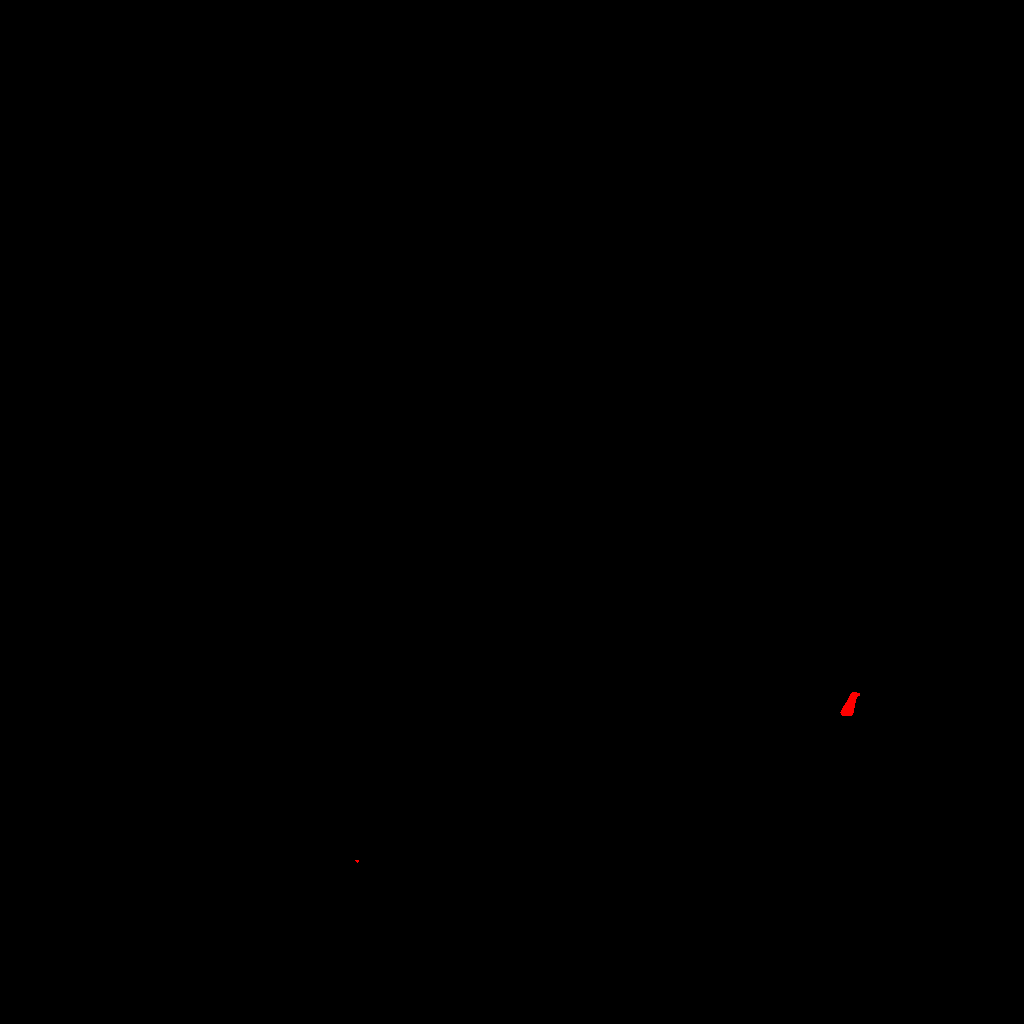} \\ 

\rotatebox{90}{\scriptsize{CHNet Vis \cite{wang2025efficient}}} & 
\includegraphics[width=.16\textwidth, height=2.21cm]{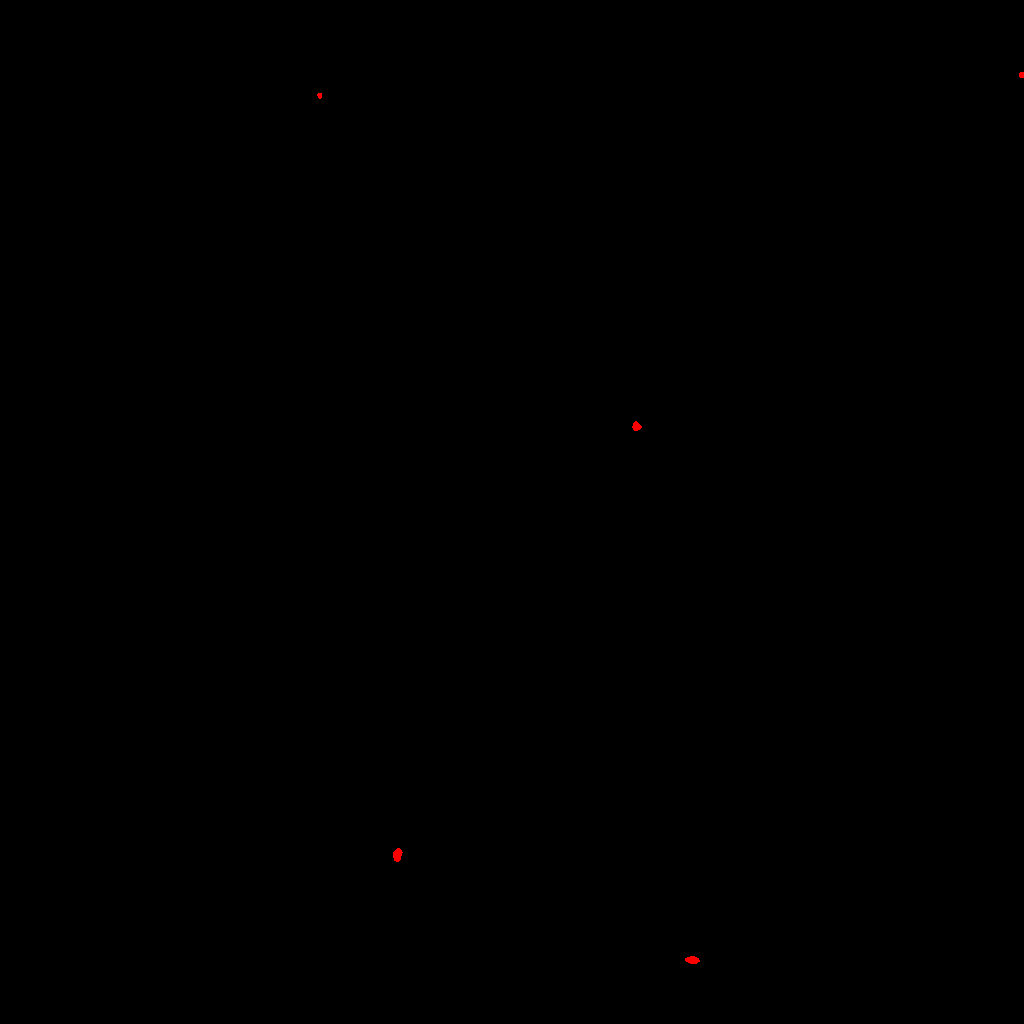} & 
\includegraphics[width=.16\textwidth, height=2.21cm]{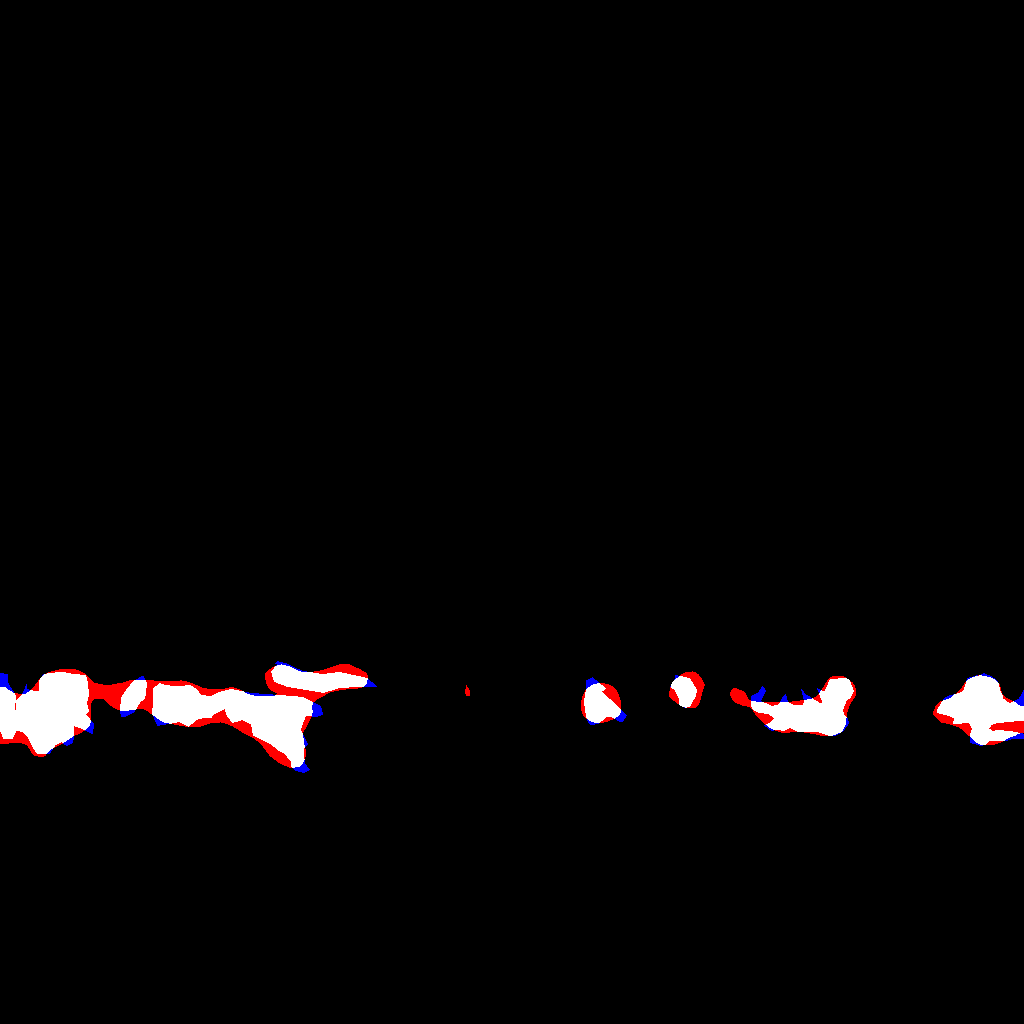} & 
\includegraphics[width=.16\textwidth, height=2.21cm]{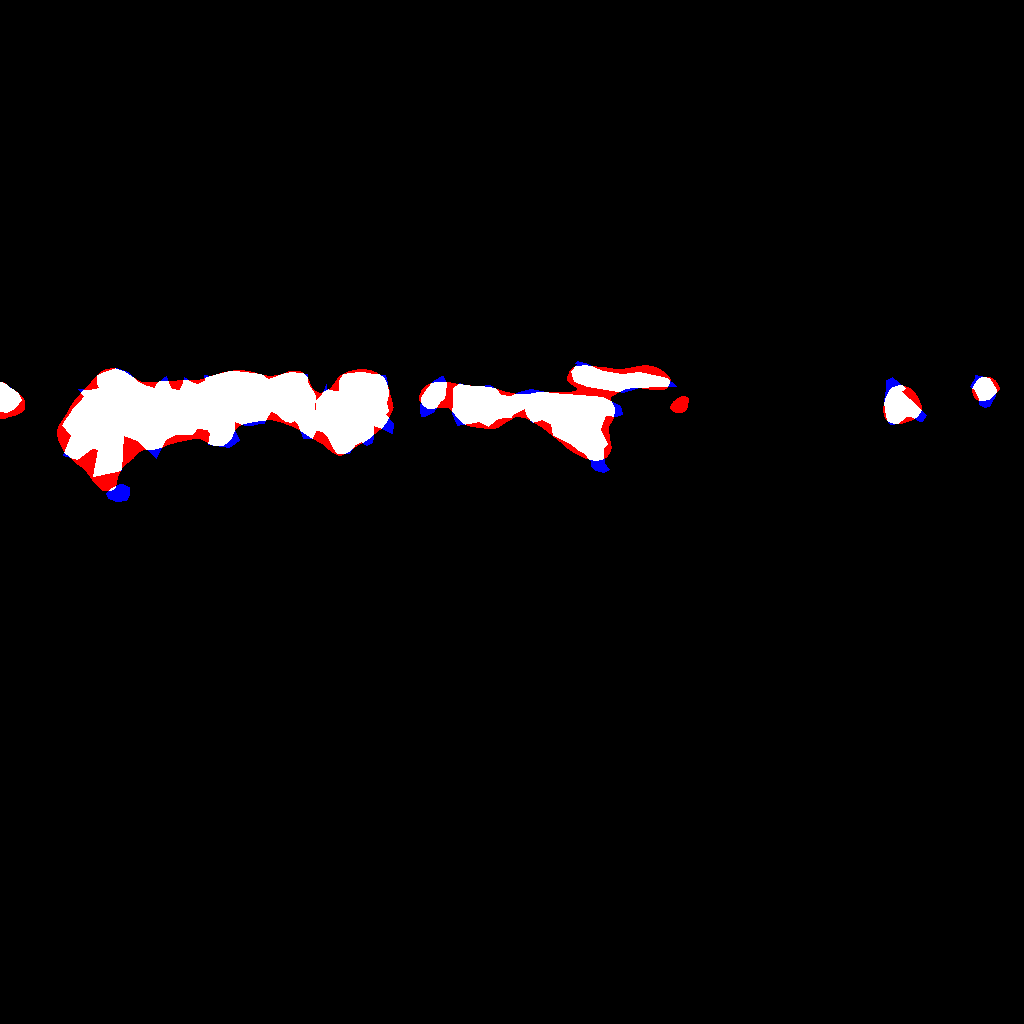} & 
\includegraphics[width=.16\textwidth, height=2.21cm]{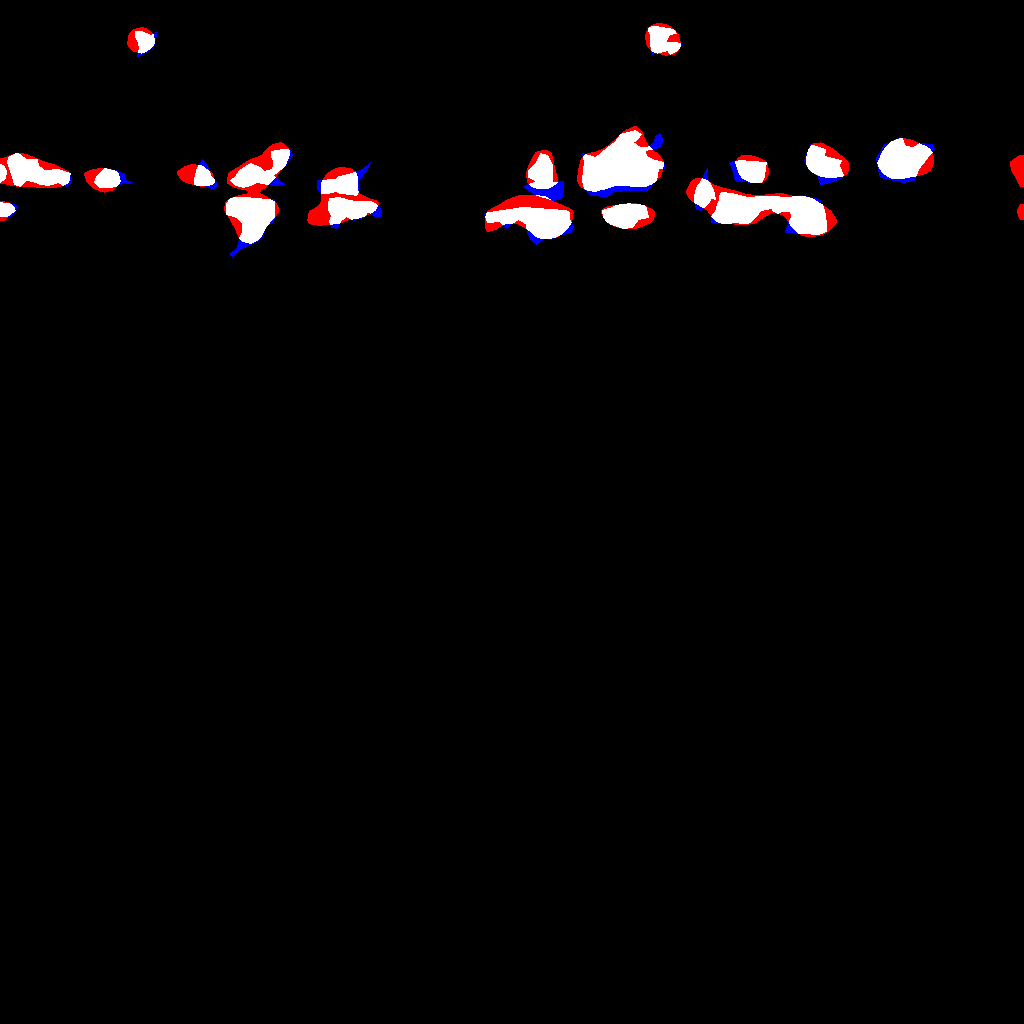} & 
\includegraphics[width=.16\textwidth, height=2.21cm]{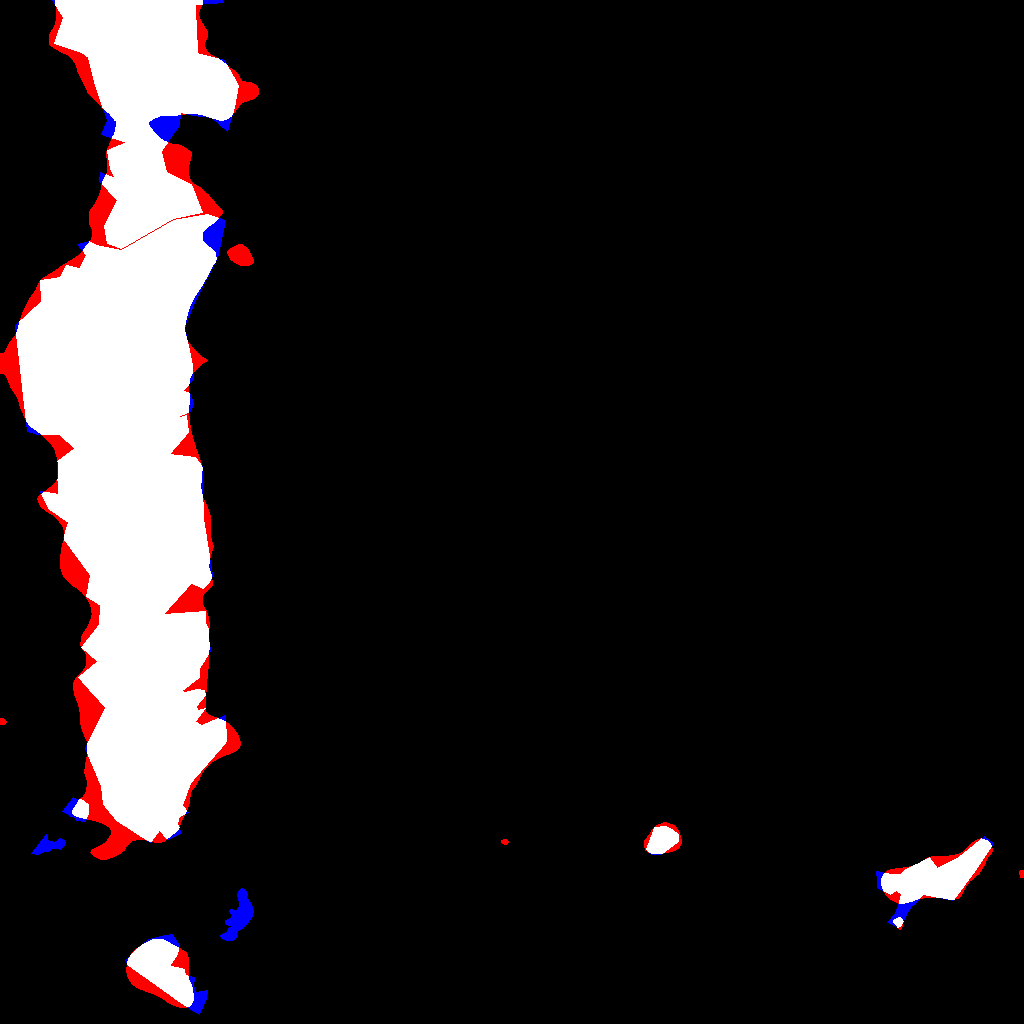} &  
\includegraphics[width=.16\textwidth, height=2.21cm]{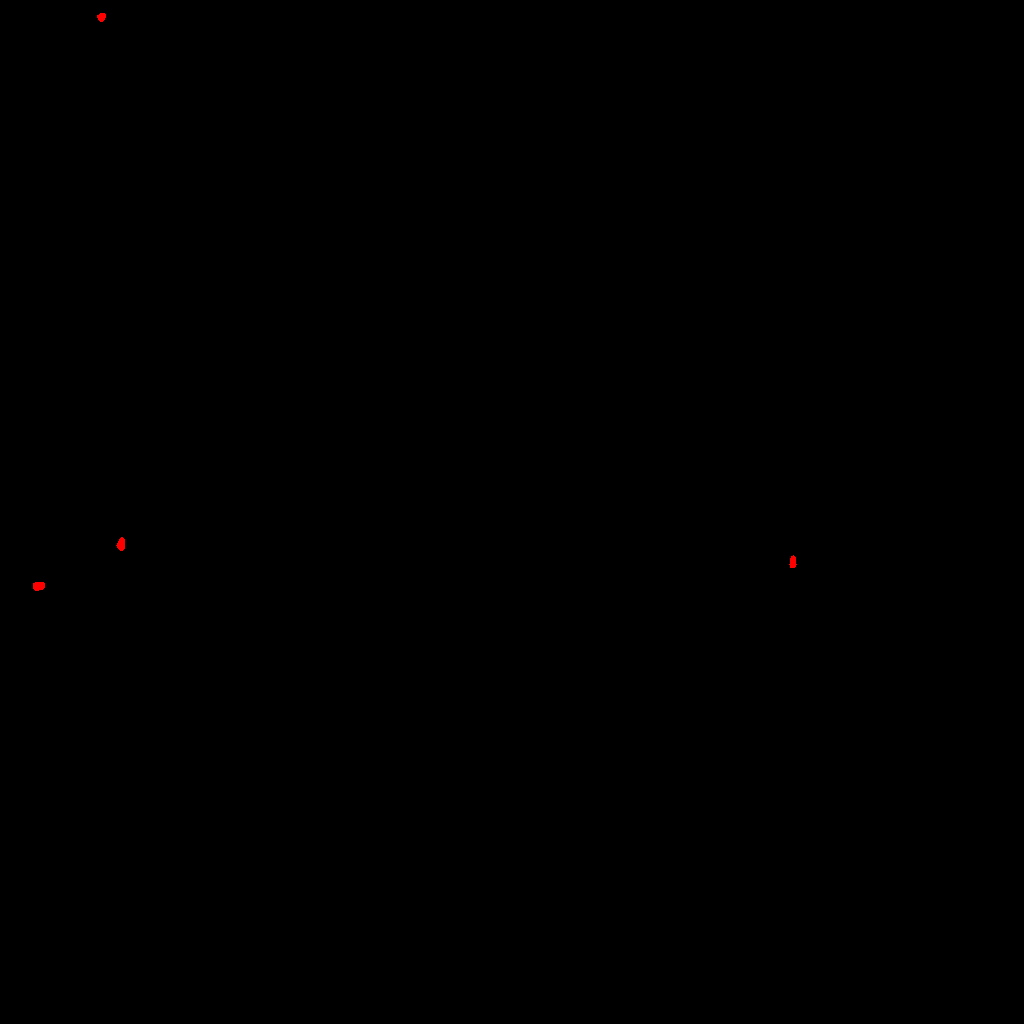} \\ 

\rotatebox{90}{\scriptsize{ARNet-v2 Vis \cite{wang2025assistedv2}}} & 
\includegraphics[width=.16\textwidth, height=2.21cm]{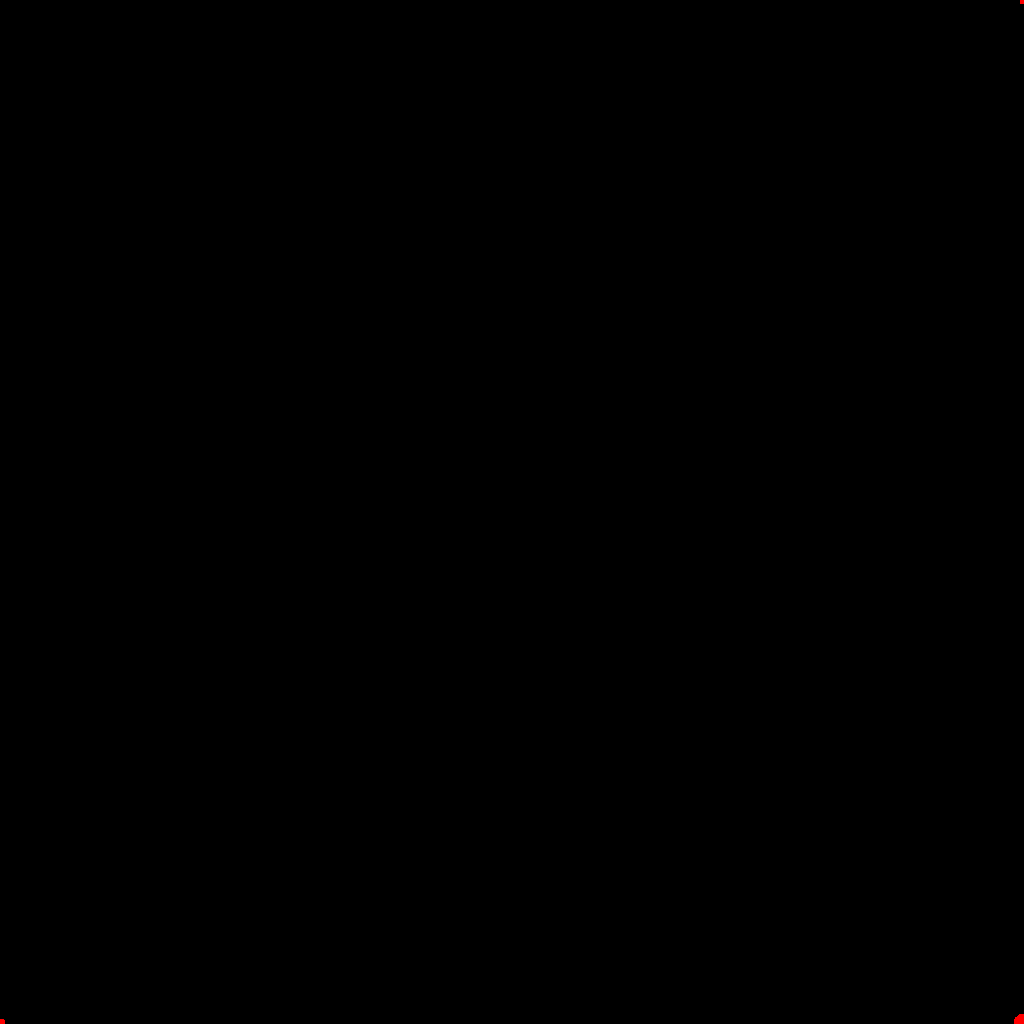} & 
\includegraphics[width=.16\textwidth, height=2.21cm]{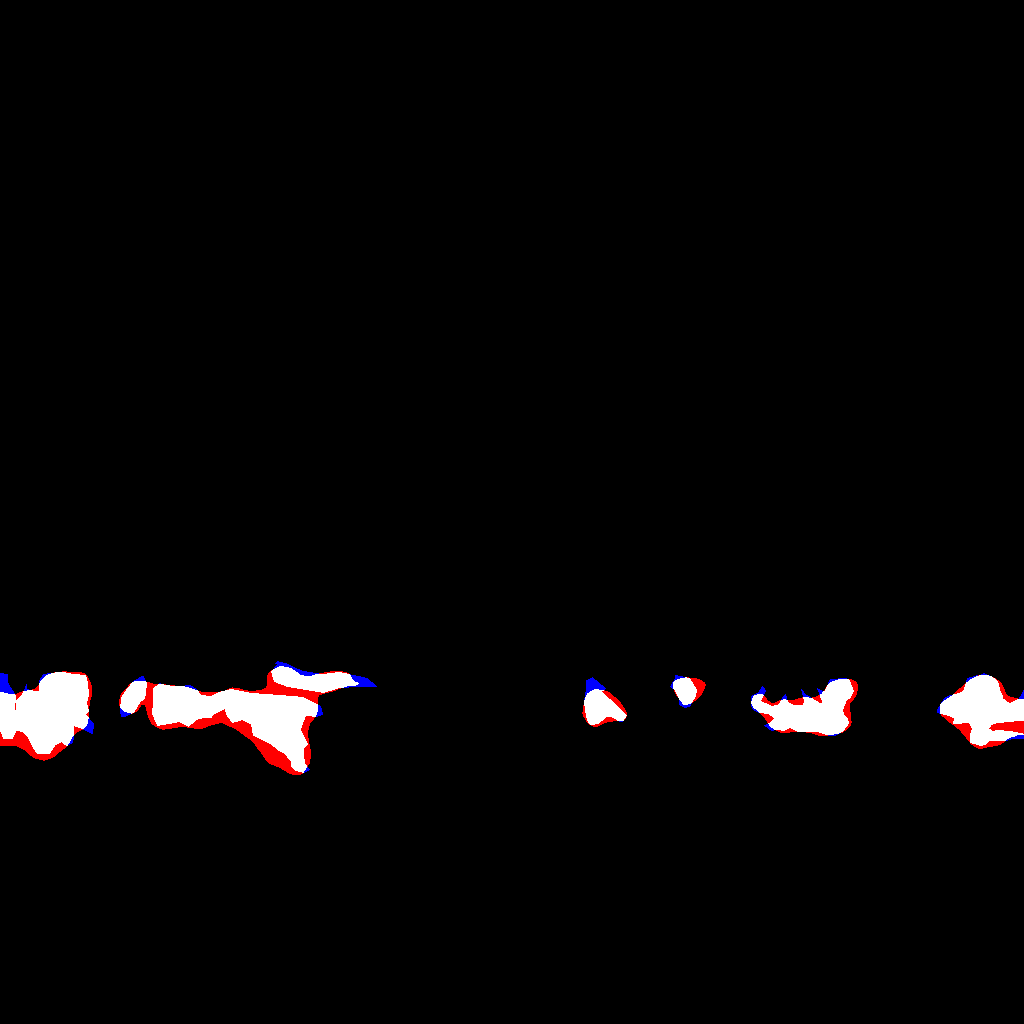} & 
\includegraphics[width=.16\textwidth, height=2.21cm]{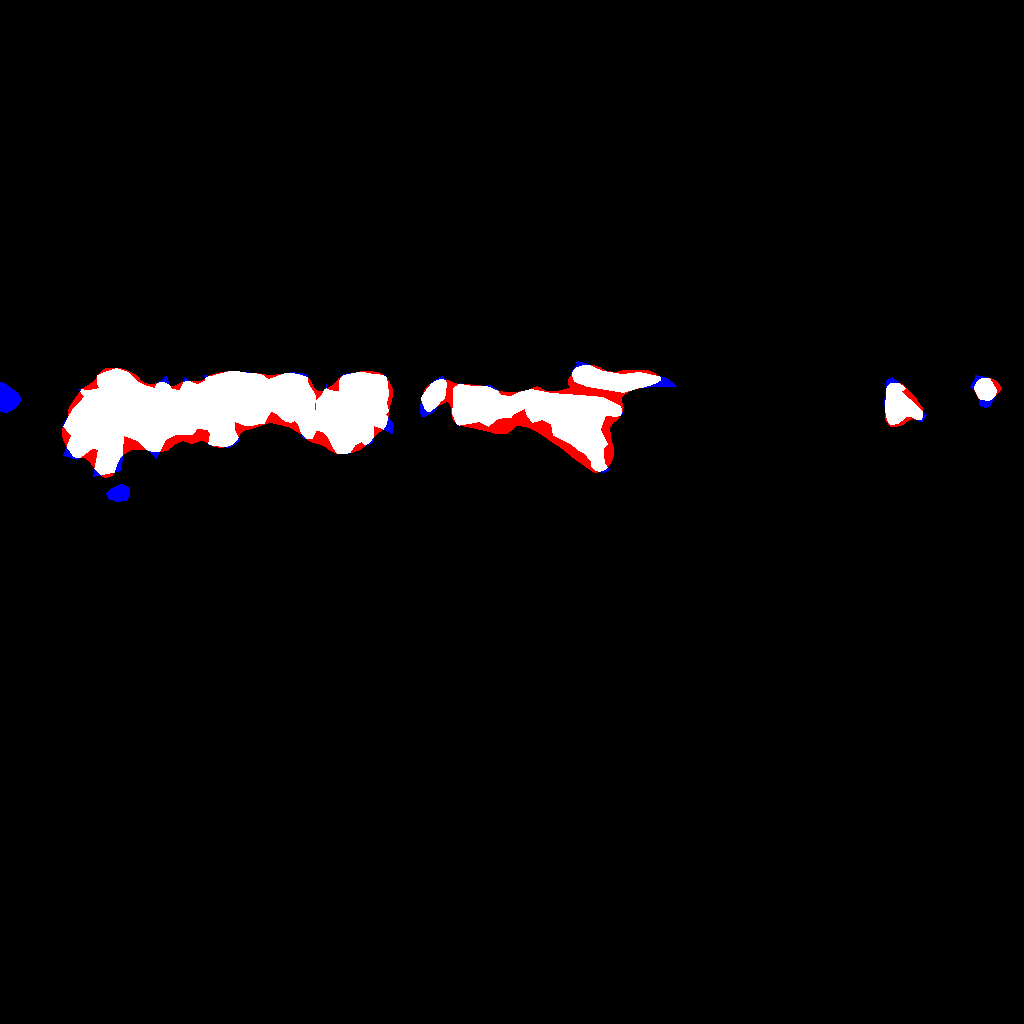} & 
\includegraphics[width=.16\textwidth, height=2.21cm]{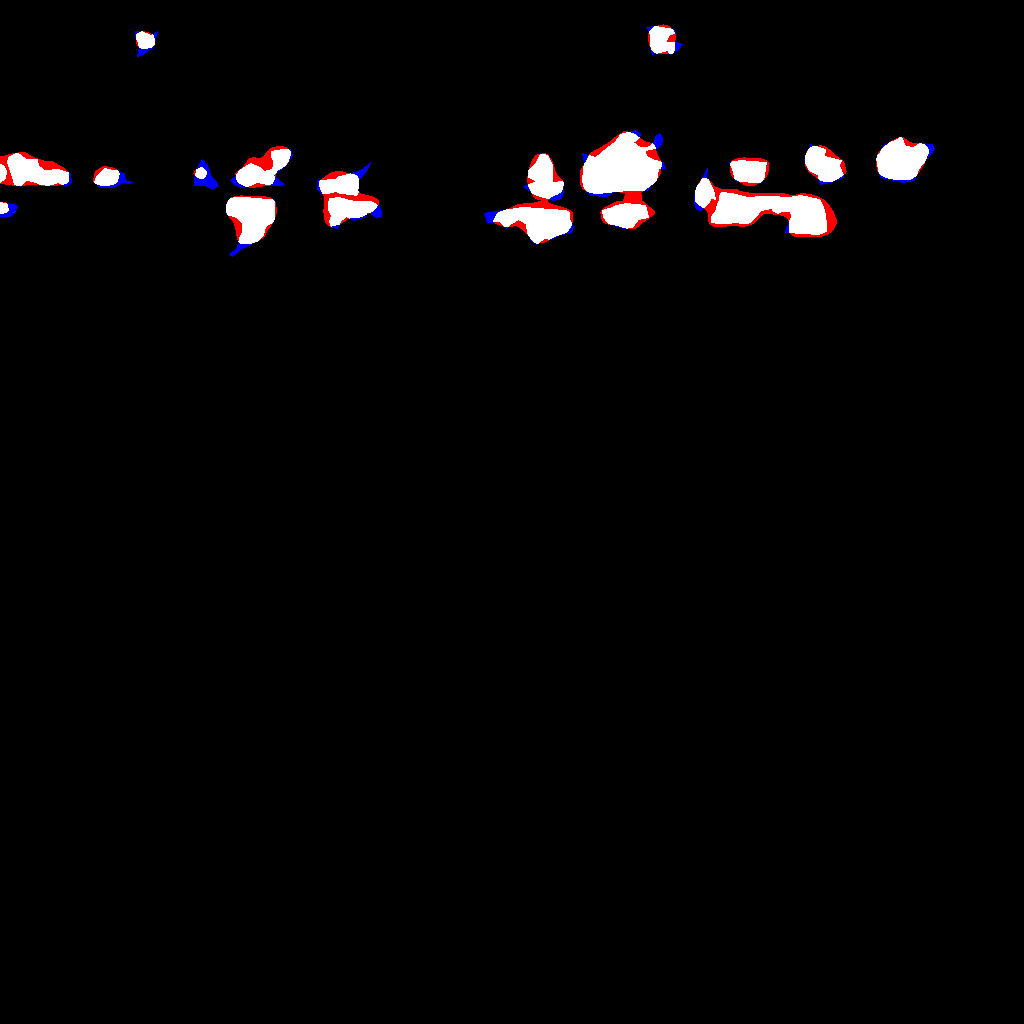} & 
\includegraphics[width=.16\textwidth, height=2.21cm]{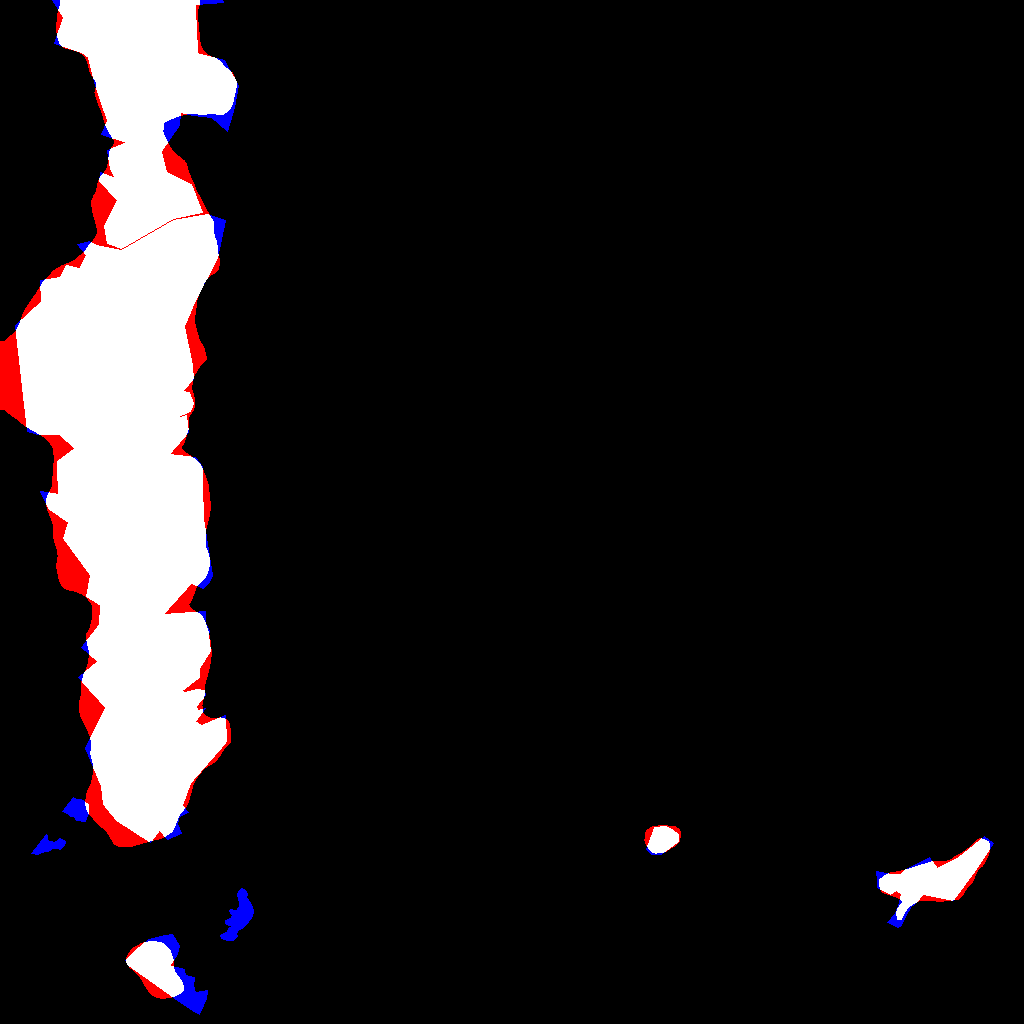} &  
\includegraphics[width=.16\textwidth, height=2.21cm]{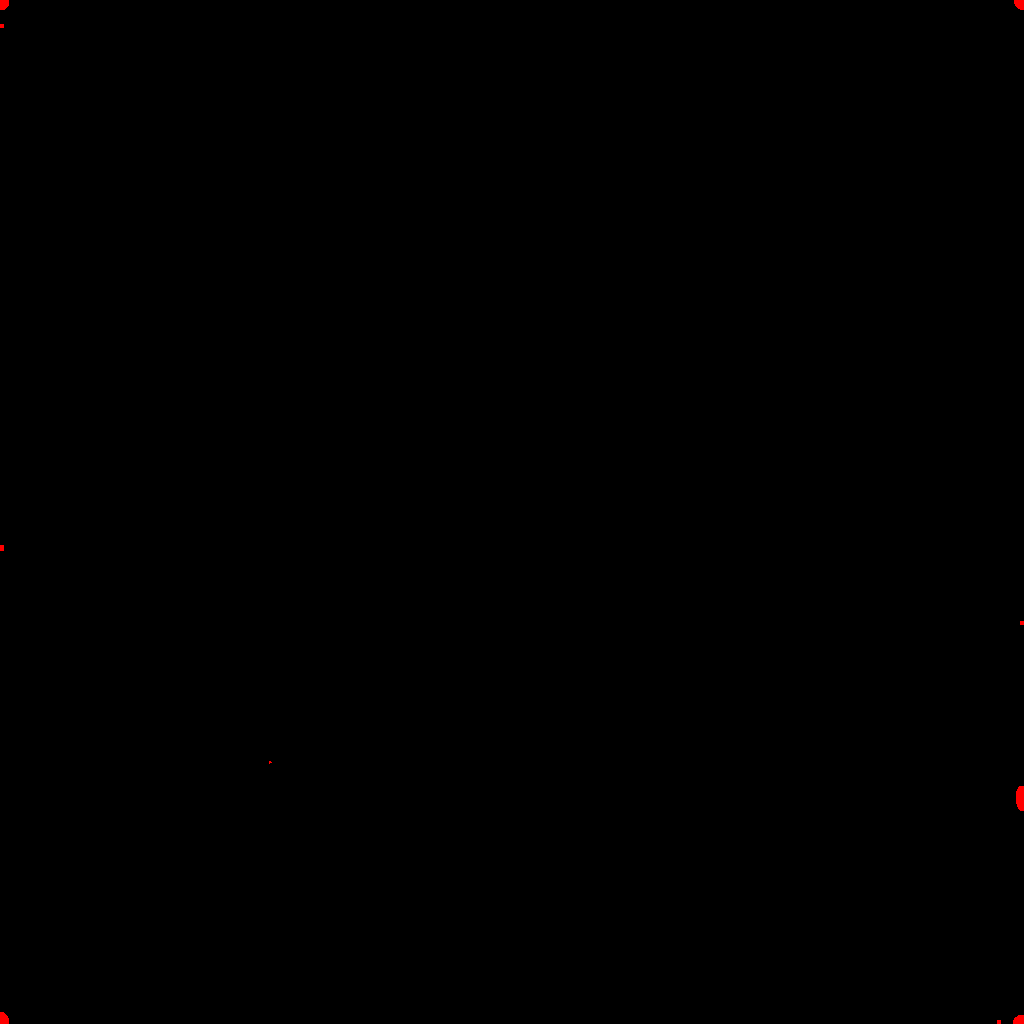} \\ 

\rotatebox{90}{\scriptsize{SWNet Vis+NIR}} & 
\includegraphics[width=.16\textwidth, height=2.21cm]{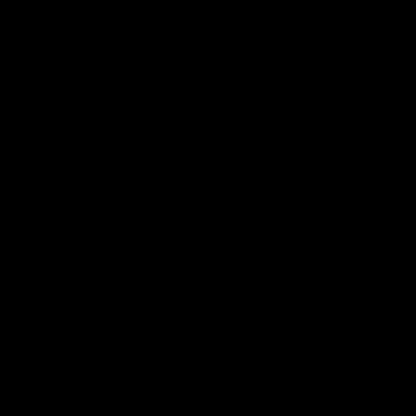} & 
\includegraphics[width=.16\textwidth, height=2.21cm]{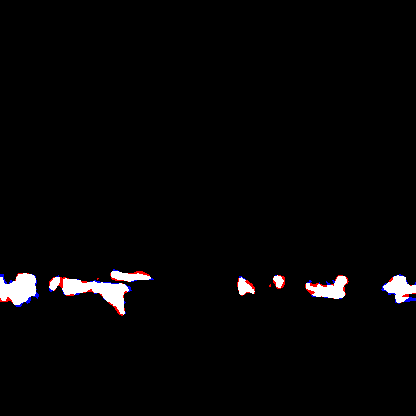} & 
\includegraphics[width=.16\textwidth, height=2.21cm]{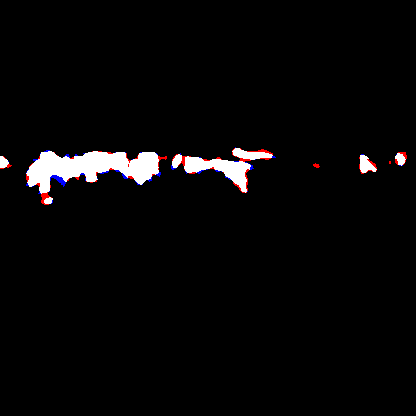} & 
\includegraphics[width=.16\textwidth, height=2.21cm]{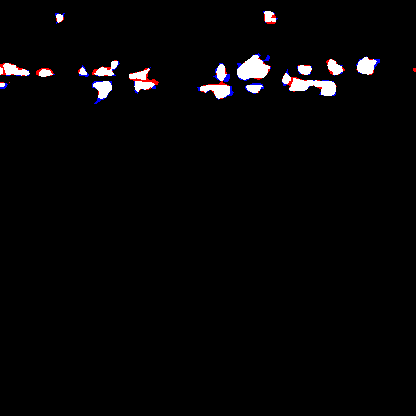} & 
\includegraphics[width=.16\textwidth, height=2.21cm]{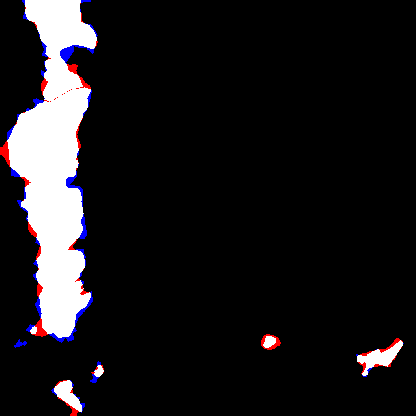} &  
\includegraphics[width=.16\textwidth, height=2.21cm]{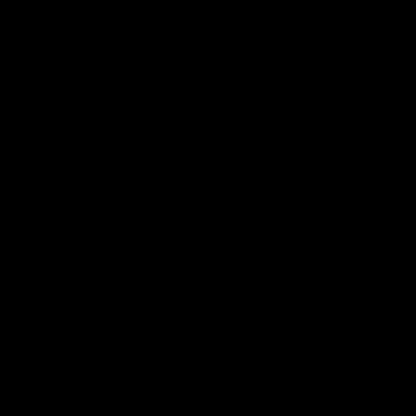} \\ 

\end{tabular}
}
\caption{Results on COD techniques that have achieved first or second place in at least one of the metrics in Table \ref{tab:results_cod_weedsbanana}. Successful matches between GT and predicted masks (white areas); False positive regions (red areas, over-segmentation); and false negative regions (blue areas, miss-segmentation).}
\label{fig:results}
\end{figure*}

\section{Experimental Results}
\label{sec:exp}
This section presents a comprehensive evaluation of the proposed SWNet against state-of-the-art (SOTA) Camouflaged Object Detection (COD) methods. We describe the quantitative performance across multiple metrics and provide a qualitative analysis of the detection results in challenging agricultural scenarios.

\subsection{Quantitative Evaluation}
The quantitative results, summarized in Table \ref{tab:results_cod_weedsbanana}, demonstrate that SWNet significantly outperforms existing SOTA methods by effectively leveraging multimodal data. When comparing single-modality baselines, our model already shows competitive results; however, the integration of Visible (Vis) and Near-Infrared (NIR) spectra provides a definitive performance leap. Specifically, SWNet (Vis+NIR) achieves a weighted F-measure ($F_{\beta}^{w}$) of 0.8767, surpassing the previous best visible-light model, ARNet, which reached 0.8131. This improvement is also reflected in the Mean Absolute Error ($M$), where SWNet achieves a record low of 0.0070, indicating superior pixel-wise precision compared to the 0.0086 achieved by ARNet-v2. Furthermore, our model reaches an S-measure ($S_{\alpha}$) of 0.8966, outperforming the specialized HitNet (0.8773) and CHNet (0.8839), which suggests that the Bimodal Gated Fusion Module is more adept at capturing the subtle structural discrepancies between weeds and crops than traditional single-stream or attention-based backbones. %Although SWNet utilizes a larger parameter count of 42.32M compared to the more compact DGNet (8.30M), the substantial gains in sensitivity and boundary accuracy justify the increased architectural complexity for high-precision agricultural applications.

\subsection{Qualitative Evaluation}
The qualitative analysis illustrated in Fig. \ref{fig:results} reinforces the numerical findings, particularly in scenarios characterized by "homochromatic blending" where target and background textures are nearly indistinguishable. While competitive models like BGNet and HitNet often suffer from significant over-segmentation (false positives) or fail to capture the complete geometry of the weed (false negatives), SWNet produces segmentation masks that closely align with the GT. The effectiveness of the Edge-Aware Refinement module is evident in the sharp transitions at object boundaries, effectively "carving" the weed out of the crop canopy where other models produce blurred or fragmented results. By utilizing the NIR spectrum to "break" the visual camouflage, SWNet maintains structural integrity in the predicted masks, even in high-clutter areas where ARNet-v2 and CHNet show visible degradation.

\begin{table*}[!h]
    \centering
    %\scriptsize
    \caption{Metric evaluation results for each COD technique on the Weeds-Banana \cite{velesaca2026unveiling} dataset, reported adding different module on proposed bimodal cross-spectral architecture. Results are presented using the metric notation defined in Sec.~\ref{subSec:metricsEval}, ``$\uparrow/\downarrow$'' indicates that larger or smaller is better. The best three performing results are highlighted using color: \First{First}, \Second{Second}, and \Third{Third} respectively.}
    \resizebox{2\columnwidth}{!}{
    \begin{tabular}{l|rrrrrrrrr}
        \toprule
        Module & $S_\alpha \uparrow$ & $F^{w}_\beta \uparrow$ & $M \downarrow$ & $E^{adp}_\phi \uparrow$ & $E^{mean}_\phi \uparrow$ & $E^{max}_\phi \uparrow$ & $F^{adp}_\beta \uparrow$ & $F^{mean}_\beta \uparrow$ & $F^{max}_\beta \uparrow$ \\
        
        \midrule
        only Edge & 0.8797 & 0.8332 & 0.0071 & 0.9467 & 0.9439 & 0.9480 & 0.8177 & 0.8179 & 0.8339 \\
        only CBAM & 0.8714 & 0.8162 & 0.0077 & 0.9414 & 0.9415 & 0.9459 & 0.7918 & 0.8008 & 0.8204 \\
        Edge + CBAM & \textbf{0.8966} & \textbf{0.8767} & \textbf{0.0070} & \textbf{0.9857} & \textbf{0.9860} & \textbf{0.9906} & \textbf{0.8493} & \textbf{0.8590} & \textbf{0.8788} \\
        \bottomrule
    \end{tabular}
    }
    \label{tab:results_ablation}
\end{table*}

\begin{figure}[!h]
\setlength\tabcolsep{0.75pt}
\centering
\scalebox{1.0}{
\begin{tabular}{ccccccccc}

\rotatebox{90}{RGB} & 
\includegraphics[width=.22\textwidth, height=2.30cm]{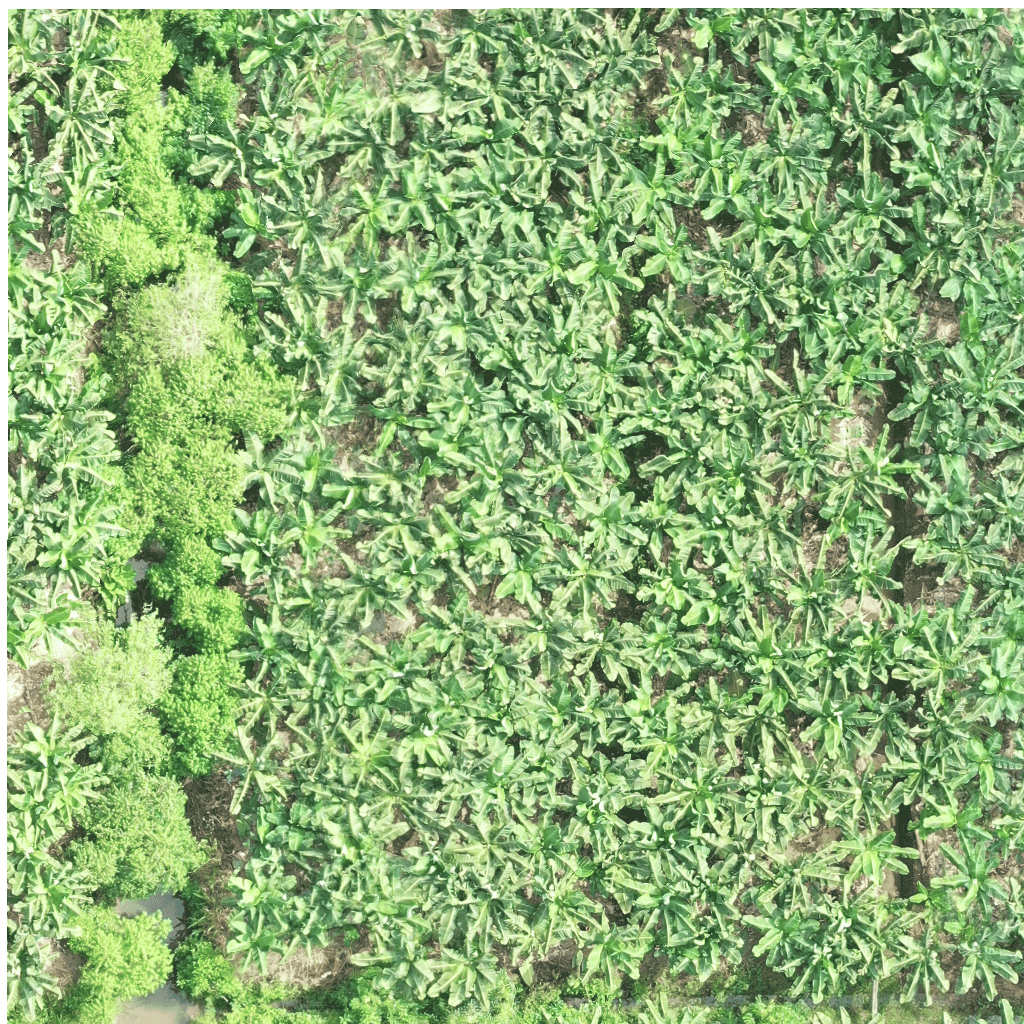} & 
\includegraphics[width=.22\textwidth, height=2.30cm]{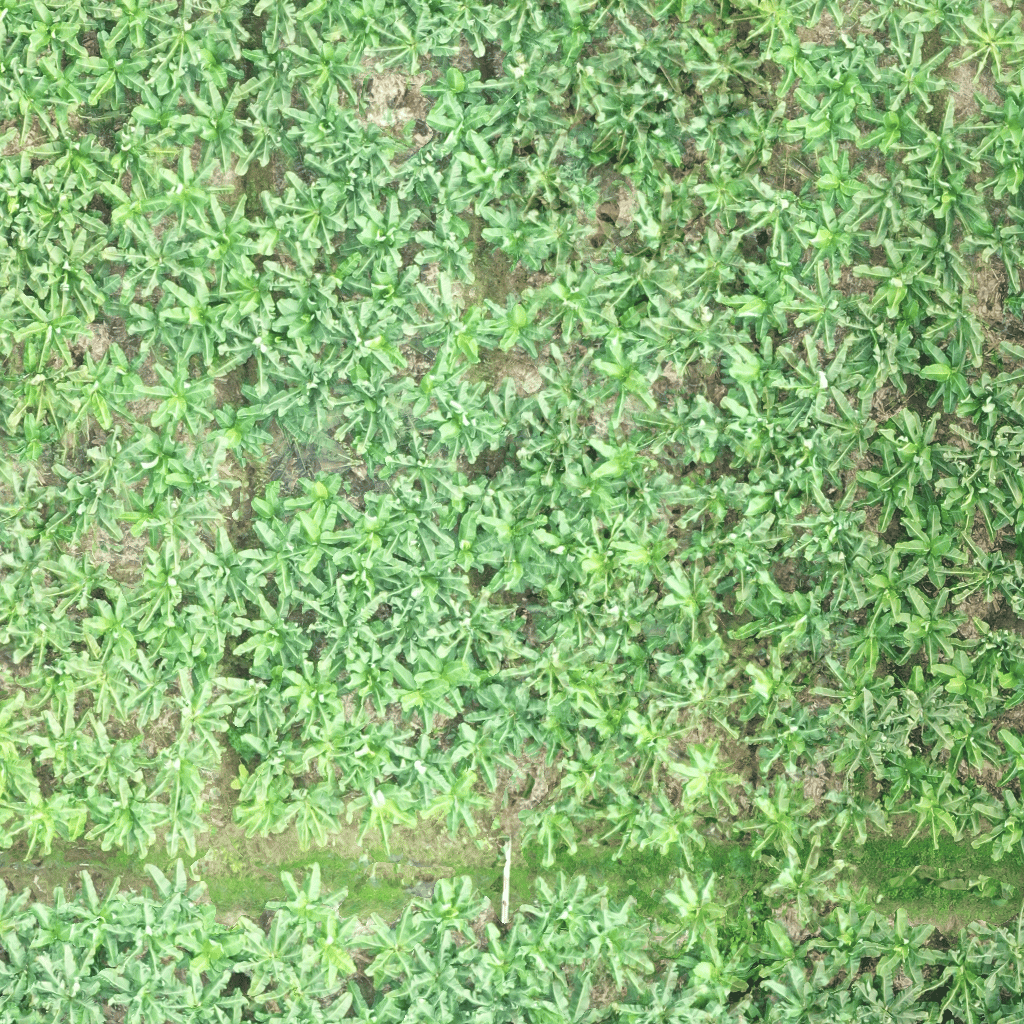} \\

\rotatebox{90}{NIR} & 
\includegraphics[width=.22\textwidth, height=2.30cm]{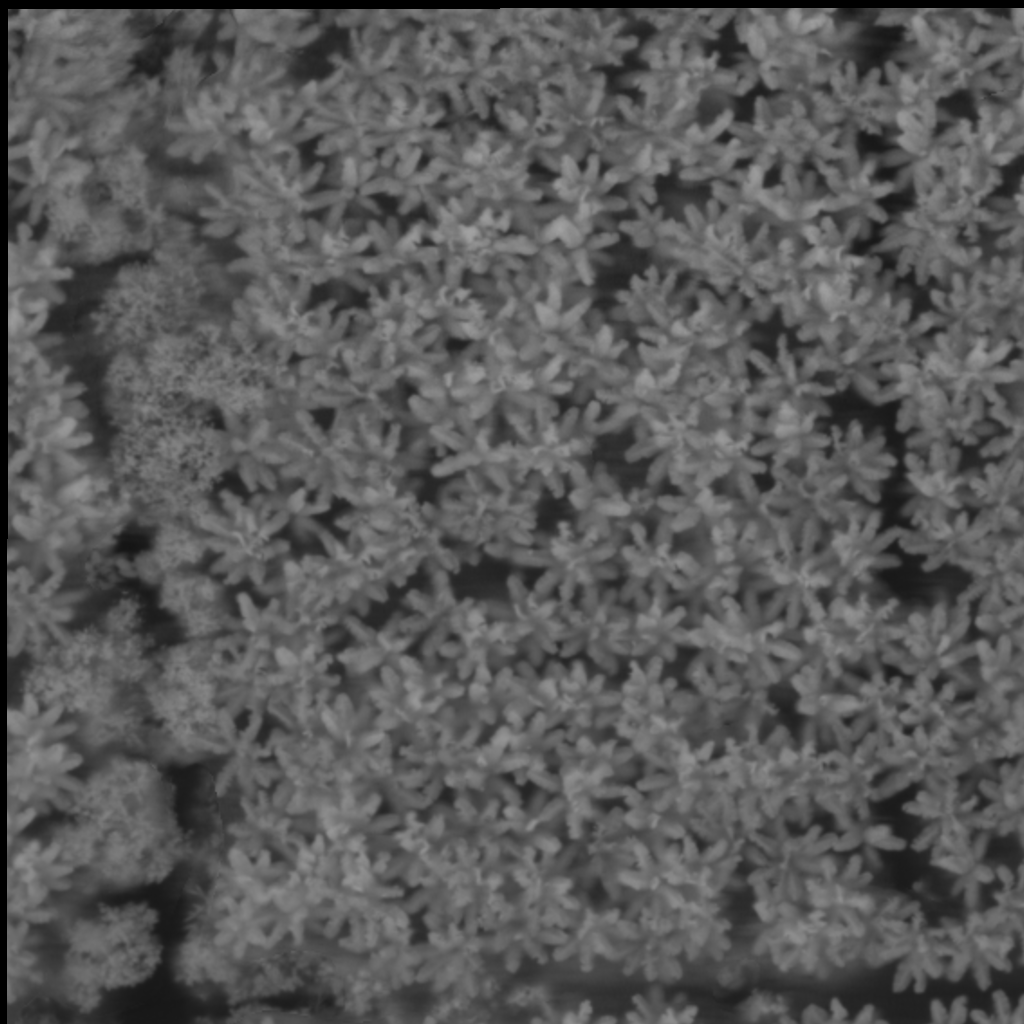} & 
\includegraphics[width=.22\textwidth, height=2.30cm]{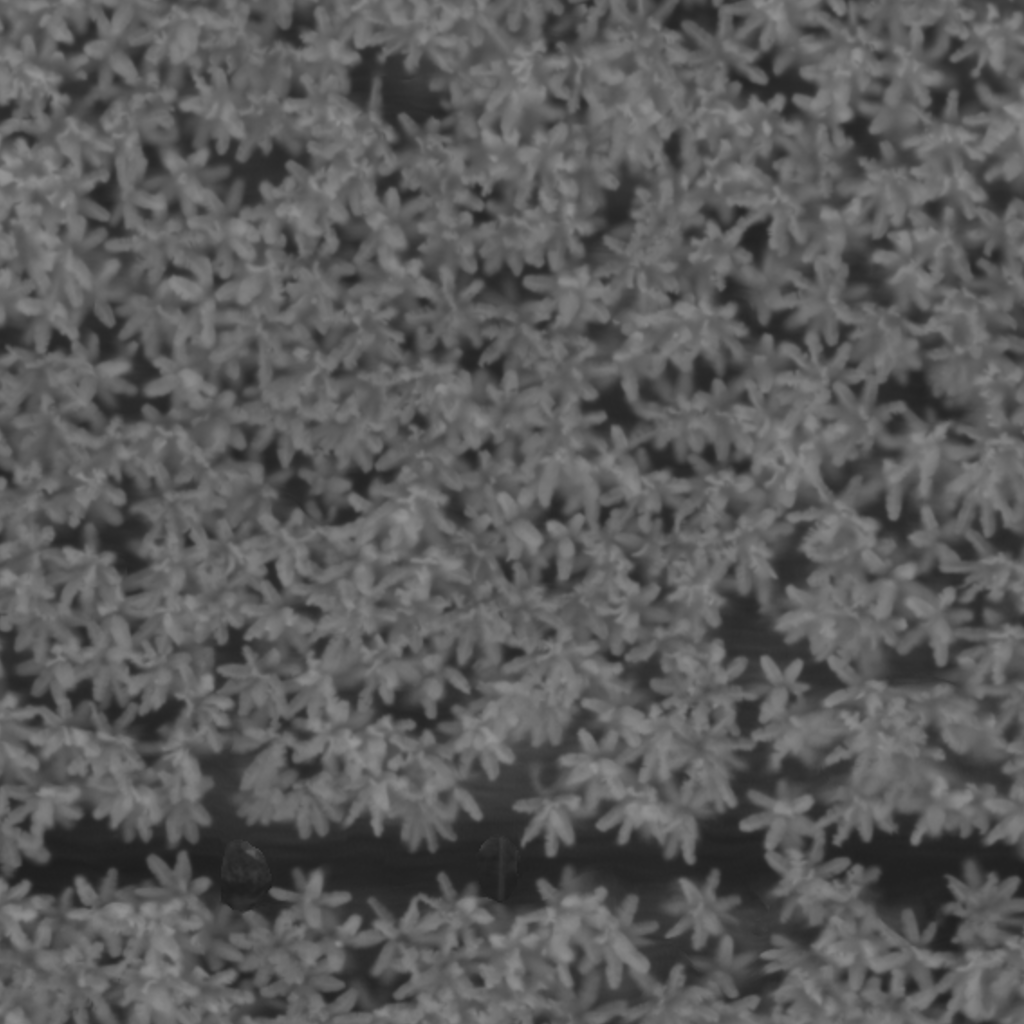} \\

\rotatebox{90}{GT} & 
\includegraphics[width=.22\textwidth, height=2.30cm]{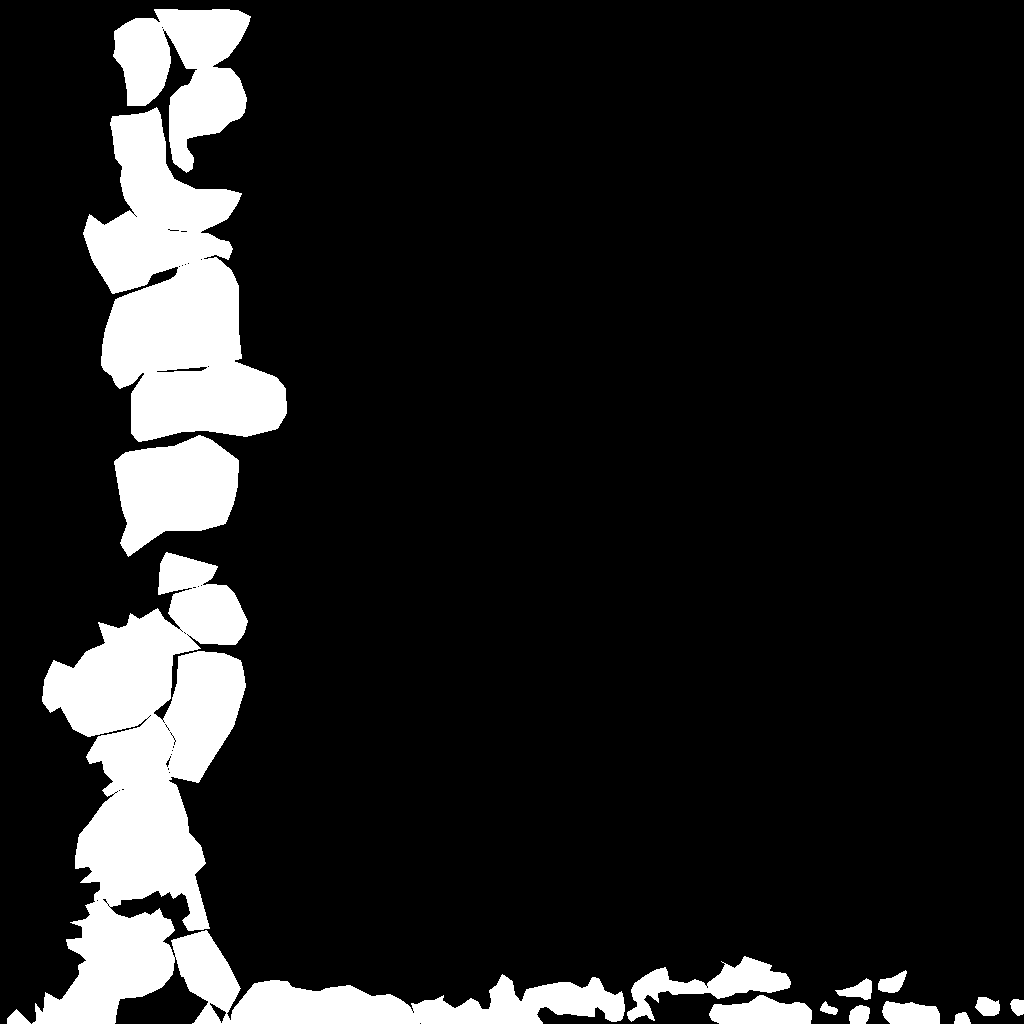} & 
\includegraphics[width=.22\textwidth, height=2.30cm]{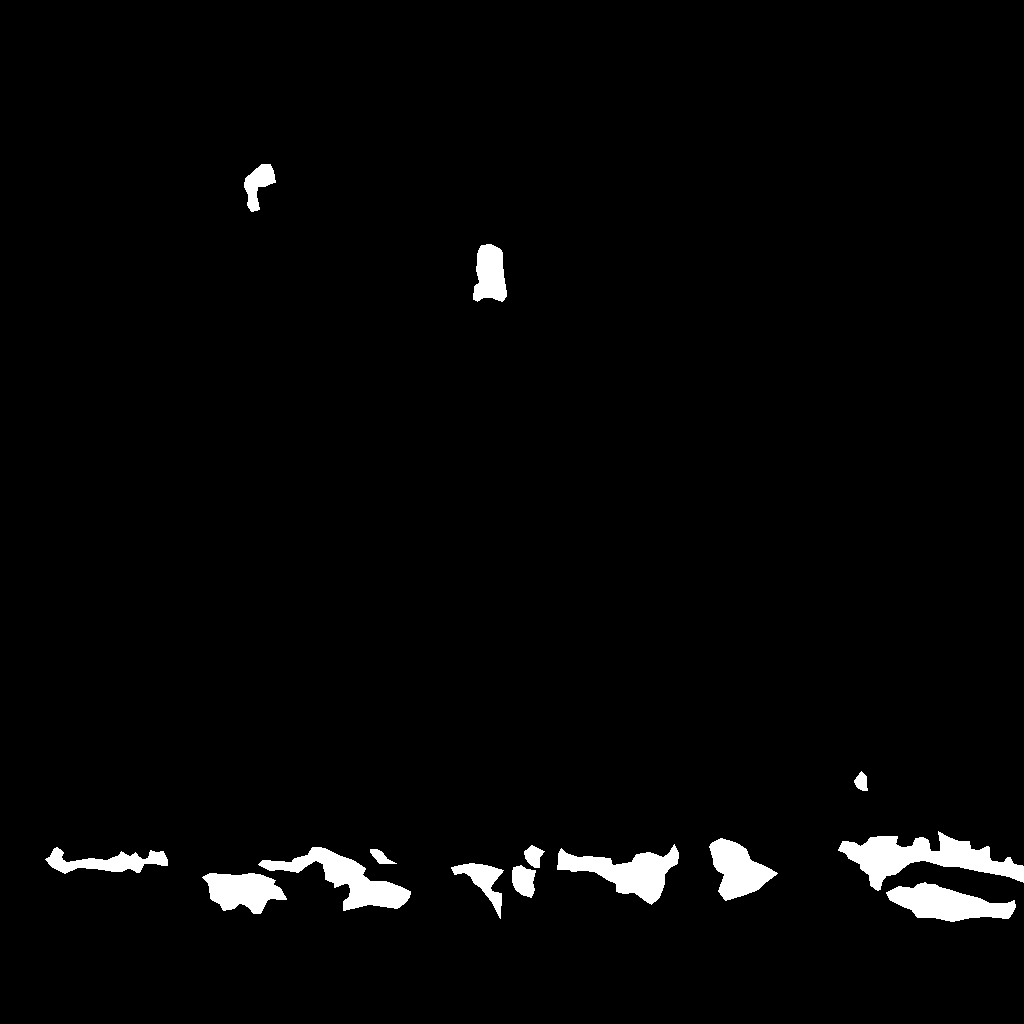} \\

\rotatebox{90}{Edge} & 
\includegraphics[width=.22\textwidth, height=2.30cm]{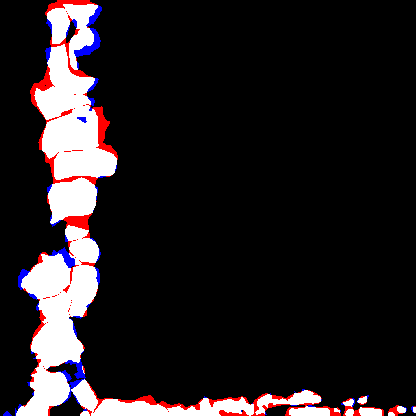} & 
\includegraphics[width=.22\textwidth, height=2.30cm]{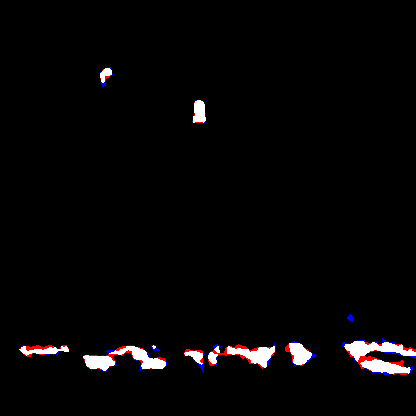} \\ 

\rotatebox{90}{CBAM} & 
\includegraphics[width=.22\textwidth, height=2.30cm]{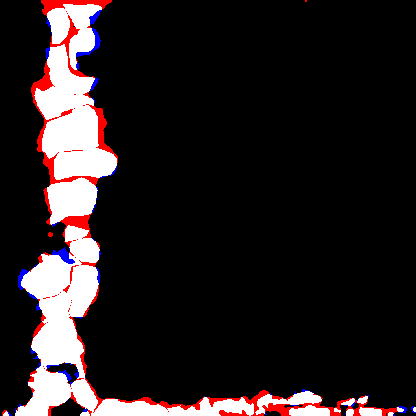} & 
\includegraphics[width=.22\textwidth, height=2.30cm]{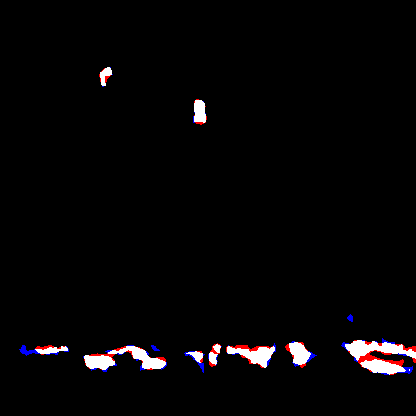} \\ 

\rotatebox{90}{Edge + CBAM} & 
\includegraphics[width=.22\textwidth, height=2.30cm]{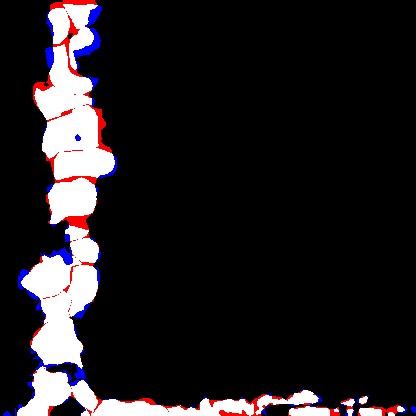} & 
\includegraphics[width=.22\textwidth, height=2.30cm]{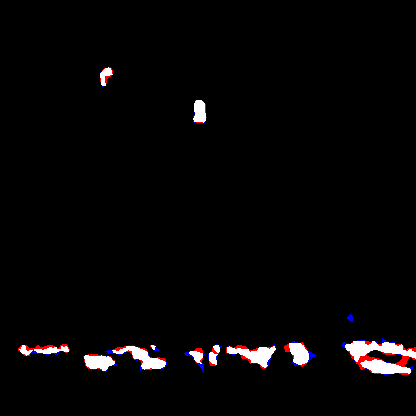} \\ 

\end{tabular}
}
\caption{Comparison of three distinct configurations: a version using only the Edge module, one using only the CBAM module, and the complete SWNet integrating both. Successful matches between GT and predicted masks (white areas); False positive regions (red areas, over-segmentation); and false negative regions (blue areas, miss-segmentation).}
\label{fig:results}
\end{figure}

\subsection{Ablation study}
To evaluate the effectiveness of the key components within the SWNet architecture, ablation experiments are conducted focusing on the two main enhancement modules: the Edge Refinement module and the CBAM. Table \ref{tab:results_ablation} compares three distinct configurations: a version utilizing only the Edge module, one utilizing only the CBAM module, and the complete SWNet integrating both. The results indicate that while the Edge module improves structural boundary definition ($S_{\alpha}$ of 0.8797) and the CBAM module enhances feature prioritization, their individual performances are surpassed by their combined integration. The full SWNet configuration achieved the highest scores across all metrics, notably reaching an $S_{\alpha}$ of 0.8966 and a substantial improvement in the weighted F-measure ($F_{\beta}^{w} = 0.8767$). These findings demonstrate that the synergy between spatial-channel attention and explicit edge refinement is essential for accurately discriminating camouflaged weeds from their surrounding crop environment, proving that both modules are critical for the network's success.

%\section{Discussion}
%\label{sec:disc}
%TODOS

\section{Conclusions}
\label{sec:conclu}
This research introduced SWNet, a robust bimodal architecture designed to address the problem of camouflaged weed detection through cross-spectral analysis. Extensive evaluations on the Weeds-Banana dataset demonstrate that the fusion of RGB and NIR data is critical for breaking the natural camouflage of invasive plants. The implementation of the Bimodal Gated Fusion Module and the Convolutional Block Attention Module (CBAM) enables the network to prioritize relevant spatial and channel-wise features while suppressing environmental noise. Additionally, the Edge-Aware Refinement branch significantly improves the delineation of object boundaries, providing high-fidelity masks even in regions with severe visual overlap. SWNet establishes a new performance benchmark, exceeding the results of specialized models such as ARNet-v2 and HitNet across multiple structural and pixel-wise metrics. These findings suggest that multimodal frameworks offer a superior alternative for precision agriculture, facilitating more accurate weed management in challenging field conditions. Future developments could explore the optimization of this architecture for real-time inference on edge computing platforms used in autonomous agricultural robotics.

\section*{Acknowledgements}
This material is based upon work supported by the Air Force Office of Scientific Research under award number FA9550-24-1-0206; and partially supported by the Grant PID2021-128945NB-I00 funded by MICIU/AEI/ 10.13039/501100011033 and by ERDF/EU and Grant PID2024-162815NB-I00 funded by MICIU/AEI/ 10.13039/501100011033 and by ERDF/EU; and by the ESPOL project ``Advancing Camouflaged Object Detection with a cost-effective Cross-Spectral vision system (ACODCS)'' (CIDIS-003-2024). The authors acknowledge the support of the Generalitat de Catalunya CERCA Program to CVC’s general activities.

{
    \small
    \bibliographystyle{ieeenat_fullname}
    \bibliography{main}
}


\end{document}